\documentclass[10pt,journal,compsoc]{IEEEtran}
\ifCLASSOPTIONcompsoc
  \usepackage[nocompress]{cite}
\else
  \usepackage{cite}
\fi

\usepackage{amsmath,amssymb,amsfonts}
\usepackage{algorithmic}
\usepackage{graphicx}
\usepackage{caption}
\usepackage{verbatim}
\usepackage[colorlinks = true]{hyperref}
\usepackage{color}
\usepackage{textcomp}
\begin{document}

\title{Going Deep in Medical Image Analysis: Concepts, Methods, Challenges and Future Directions}

\author{Fouzia~Altaf, Syed~M.~S.~Islam, Naveed~Akhtar, Naeem~K.~Janjua}




\IEEEtitleabstractindextext{
\begin{abstract}
Medical Image Analysis is currently experiencing a paradigm shift due to Deep Learning. This technology has recently attracted so much interest of the Medical Imaging community that it led to a specialized conference in `Medical Imaging with Deep Learning' in the year 2018.   
This article surveys the recent developments in this direction, and provides a critical review of the related  major aspects. 
We organize the reviewed literature according to the underlying Pattern Recognition tasks, and further sub-categorize it following a taxonomy based on human anatomy. 
This article does not assume prior knowledge of Deep Learning and makes a significant contribution in explaining the core Deep Learning concepts to the non-experts in the  Medical community.
Unique to this study is the Computer Vision/Machine Learning perspective taken on the advances of Deep Learning in Medical Imaging.
This enables us to single out `lack of appropriately annotated large-scale datasets' as the core challenge (among other challenges) in this research direction.
We draw on the insights from the sister research fields of Computer Vision, Pattern Recognition and Machine Learning etc.; where the techniques of dealing with such challenges have already matured, to provide promising directions for the Medical Imaging community to fully harness Deep Learning in the future.

\end{abstract}

}



\maketitle

\section{Introduction} 
\label{sec:introduction}
Deep Learning (DL)~\cite{lecun2015deep} is a major contributor of the contemporary rise of Artificial Intelligence in nearly all walks of life. This is a direct consequence of the recent   breakthroughs resulting from its application across a wide variety of scientific fields; including Computer Vision~\cite{NaveedSurvey}, Natural Language Processing~\cite{sutskever2014sequence}, Particle Physics~\cite{ciodaro2012online}, DNA analysis~\cite{xiong2015human}, brain circuits studies~\cite{helmstaedter2013connectomic},  and chemical structure analysis~\cite{ma2015deep} etc. Very recently, it has also attracted a notable interest of  researchers in Medical Imaging, holding great promises for the future of this  field.

\begin{figure*}[t]
\centering
\includegraphics[width=5.5in]{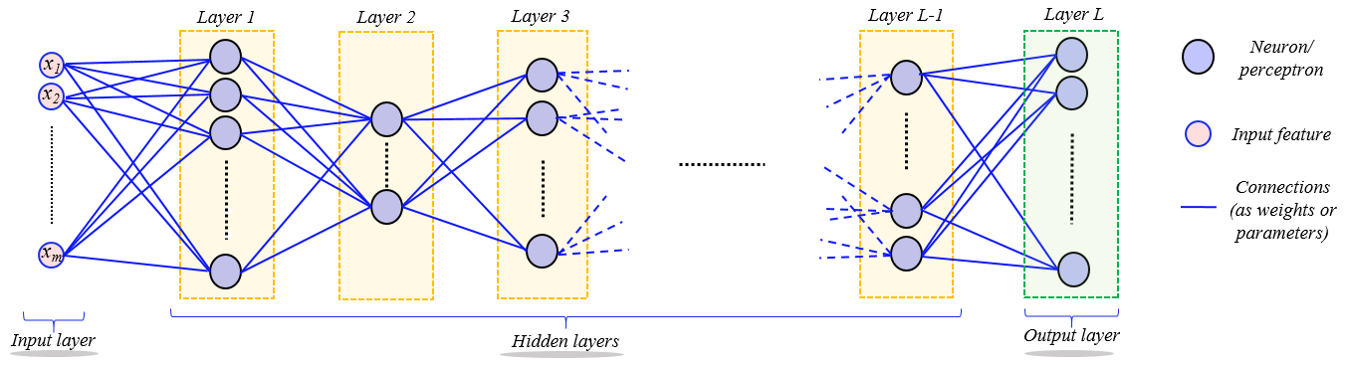}
\caption{Illustration of a deep  neural network: The network consists of multiple layers of neurons/perceptrons that are connected in an inter-layer fashion. The neurons compute non-linear/linear transformations of their inputs. Each feature of the input signal is weighted by the network parameters and processed by the neuron layers hierarchically. A network connection performs the weighting operation. 
The strengths of the connections (i.e. network parameter values) are learned using  training data.
 The network layers not seen by the input and output signals are often termed  `hidden' layers.  }
\label{fig:ANN}
\end{figure*}

The DL framework allows machines to learn very complex mathematical models for data representation, that can subsequently be used to perform accurate data analysis. These models hierarchically compute  non-linear and/or linear functions of the input data that is weighted by the model parameters. Treating these functions as data processing `layers', the hierarchical use of a large number of such layers  also inspires the name `Deep' Learning. The common  goal of DL  methods is to iteratively learn the parameters of the  computational model using a training data set such that the model gradually gets better in performing a desired task, e.g. classification; over that data under a specified metric. The computational model itself generally takes the form of an Artificial Neural Network (ANN)~\cite{schalkoff1997artificial} that consists of multiple layers of neurons/perceptrons~\cite{rosenblatt1961principles} - basic computational blocks, whereas its parameters (a.k.a.~network weights) specify the strength of the connections between the neurons of different layers. We depict a deep neural network in Fig.~\ref{fig:ANN} for illustration.

Once trained for a particular task, the DL models are also able to perform the same task accurately  using a variety of previously unseen data (i.e.~testing data). This strong generalization ability of DL currently  makes it stand out of the other Machine Learning techniques. 
Learning of the parameters of a deep model is carried out with the help of  back-propagation strategy~\cite{rumelhart1986learning} that enables some form of the popular Gradient Descent technique~\cite{kingma2014adam}, \cite{qian1999momentum} to iteratively arrive at the desired parameter values. Updating the model parameters using the complete training data \textit{once} is known as a single epoch of network/model training. Contemporary DL models are normally trained for hundreds of epochs before they can be deployed.

Although the origins of Deep Learning can be traced back to 1940s~\cite{goodfellow2016deep}, the sudden  recent rise in its utilization for solving complex problems of the modern era results from three major phenomena. (1)~Availability of large amount of training data: With ubiquitous digitization of information in recent times, very large amount of data is available to train complex computational models. Deep Learning has an intrinsic ability to model complex functions by simply stacking multiple layers of its basic computational blocks. Hence, it is a convenient choice for dealing  with hard problems.
Interestingly, this ability of deep models is known for over few decades now~\cite{beale1990neural}. However, the bottleneck of relatively smaller training data sets had restricted the utility of Deep Learning until recently. (2)~Availability of powerful computational resources: Learning  complex functions over large amount of data  results in  immense computational requirements. Related research communities are able to fulfill such requirements only recently. (3)~Availability of public libraries implementing DL algorithms: There is a growing recent trend in different research communities to publish the  source codes on public platforms. Easy public access to DL algorithm implementations has exploded the use of this technique in many application domains.  

The field of Medical Imaging has been exploiting Machine Learning since 1960s~\cite{becker1964digital}, \cite{lodwick1963coding}. However, the first notable  contributions that relate to modern Deep Learning techniques appeared in the  Medical Imaging  literature in  1990s~\cite{wu1993artificial, lo1995artificial, sahiner1996classification, chan1995computer, zhang1996improved}.  
The relatedness of these methods to contemporary DL  comes in the form of using ANNs to accomplish Medical Imaging tasks.
Nevertheless, restricted by the amount of training data and computational resources, these works trained networks that were only two to three layers deep. This is no longer considered `deep' in the modern era.
The number of layers in the contemporary DL  models  generally ranges from a dozen to over one hundred~\cite{he2016deep}.
In the context of image analysis, such models have mostly originated in  Computer Vision literature~\cite{NaveedSurvey}.

The field of Computer Vision closely relates to Medical Imaging in analyzing digital images. 
Medical Imaging has a long   tradition of profiting from the findings in Computer Vision. In 2012~\cite{krizhevsky2012imagenet}, DL provided a major breakthrough in Computer Vision by performing a very hard image classification task with remarkable accuracy.
Since then, the Computer Vision community has gradually shifted its main focus to DL. Consequently, Medical Imaging literature also started witnessing methods exploiting deep neural networks in around 2013, and now such methods are appearing at an ever increasing rate. Sahiner et al.~\cite{sahiner2018deep} noted  that the peer-reviewed publications that employed DL for radiological images tripled from 2016 ($\sim$100) to 2017 ($\sim$300), whereas the first quarter of 2018 alone saw over 100 such publications. 
Similarly,  the main stream Medical Imaging conference, i.e.~International Conference on `Medical Image Computing and Computer Assisted Intervention' (MICCAI) published over 120 papers in its main proceedings in 2018 that employed Deep Learning for Medical Image Analysis tasks.

The large inflow of contributions exploiting  DL in Medical Imaging has also given rise to a specialized venue in the form of  International Conference on
`Medical Imaging with Deep Learning' (MIDL) in 2018\footnote{\url{https://midl.amsterdam/}} that published 82 contributions in this direction. Among other published papers in 2018, our literature review also includes notable  contributions from MICCAI and MIDL 2018.
We note that few review articles~\cite{sahiner2018deep, litjens2017survey, ker2018deep} for closely related research directions also exist at the time of this publication.
Whereas those articles collectively provide a comprehensive overview of the methods exploiting deep neural networks for medical tasks until the year 2017, none of them sees the existing Medical Imaging literature through the lens of Computer Vision and Machine Learning. Consequently, they also fall short in elaborating on the root causes of the challenges faced by Deep Learning in Medical Imaging. Moreover, limited by their narrower perspective, they also do not provide insights into leveraging the findings in other fields for addressing these challenges.    

In this paper, we provide a comprehensive review of the recent DL techniques in Medical Imaging, focusing mainly on the very recent methods published in 2018. We categorize these techniques under different pattern recognition  tasks and further sub-categorize them following a taxonomy based on human anatomy. 
Analyzing the reviewed literature, we establish `lack of appropriately annotated large-scale datasets' for Medical Imaging tasks as the fundamental challenge (among other challenges) in fully exploiting Deep Learning for those tasks. We then draw on the literature of Computer Vision, Pattern Recognition and Machine Learning in general; to provide guidelines to deal with this and other challenges in Medical Image Analysis using Deep Learning. This review also touches upon the available public datasets to train DL models for the medical imaging tasks. 
Considering the lack of in-depth comprehension of Deep Learning framework by the  broader Medical community, this article also provides the understanding of the core  technical concepts related to DL at an appropriate level. This is an intentional contribution of this work.

The remaining article is organized as follows. In Section~\ref{sec:Bg}, we present the core Deep Learning concepts for the Medical community in an intuitive manner. The main body of the reviewed literature is presented in Section~\ref{sec:DLMethods}. We touch upon the public data repositories for Medical Imaging in Section~\ref{sec:Dataset}. In Section~\ref{sec:chal}, we highlight the major challenges faced by Deep Learning in Medical Image Analysis. Recommendations for dealing with these challenges are discussed in Section~\ref{sec:FD} as future directions. The article concludes in Section~\ref{sec:conc}.




\section{Background Concepts}
\label{sec:Bg}
In this Section, we first briefly introduce the broader types of Machine Learning techniques and then focus on the Deep Learning framework. Machine Learning methods are boradly categorized as \textit{supervised} or  \textit{unsupervised} based on the training data used to learn the computational models. Deep Learning based methods can fall in any of these categories. 


\paragraph{Supervised Learning}
\label{p:SL}
In supervised  learning, it is  assumed that the training data is available in the form of pairs $({\boldsymbol x}, {\boldsymbol y})$, where $\boldsymbol x \in \mathbb R^m$ is a {\it training example}, and ${\boldsymbol y}$ is its \textit{label}. The training examples generally belong to different, say $C$ \textit{classes} of data.  In that case, ${\boldsymbol y}$ is often represented as a binary vector living in $\mathbb R^{C}$, such that its $c^{\text{th}}$ coefficient is `$1$' if $\boldsymbol x$ belongs to the $c^{\text{th}}$ class, whereas all other coefficients are zero. A typical task for supervised learning is to find a computational model $\mathcal M$ with the help of training data such that it is also able to correctly predict the labels of data samples that it had not seen previously during training. The unseen data samples are termed \textit{test/testing samples} in the Machine Learning parlance. To learn a model that can perform successful  classification of the test samples, we can formulate our learning problem as estimation of the parameters $\boldsymbol \Theta$ of our model that minimizes a specific  \textit{loss}  $\mathcal{L} (\boldsymbol y, \boldsymbol{\hat y})$, where $\boldsymbol{ \hat y}$ is the label vector \textit{predicted} by the model for a given test sample. As can be guessed, the loss is defined such that it has a small value only if the learned parametric model is able to predict the correct label of the  data sample.  
Whereas the model loss has its scope limited to only a single data sample, we define a \textit{cost} for the complete training data. The cost of a model is simply the Expected value of the losses computed for the individual data samples. 
Deep Learning allows us to learn model  parameters $\boldsymbol \Theta$ that are able to achieve very low cost over very large data sets.

Whereas classification generally aims at learning computational models that map input signals to discrete output values, i.e. class labels.  It is also possible to learn models that can map training examples to continuous output values. In that case, $\boldsymbol y$ is typically a real number scalar or vector. An example of such task is to learn a model that can predict the probability of a tumor being benign or malignant. In Machine Learning, such a task is seen as a {\it regression problem}. Similar to the classification problems, Deep Learning has been able to demonstrate excellent performance in learning computational models for the regression problem.

\paragraph{Unsupervised Learning}
\label{p:UL}
Whereas supervised learning assumes that the training data also provides labels of the samples;  {\it unsupervised} learning  assume that sample labels are not available. In that case, the typical task of a computational model is to {\it cluster} the data samples into different groups based on the similarities of their intrinsic characteristics - e.g.~clustering pixels of color images based on their RGB  values. Similar to the supervised learning tasks, models for unsupervised learning tasks can also take advantage of minimizing a loss function.
In the context of Deep Learning, this loss function is normally designed such that the model learns an accurate mapping of an input signal to itself. Once the  mapping is learned, the model is used to compute compact representations of data samples that cluster well. 
Deep Learning framework has also been found very effective for unsupervised learning.

Along with supervised and unsupervised learning, other machine learning types include \textit{semi-supervised learning} and \textit{reinforcement learning}. Informally, semi-supervised learning  computes models using the training data that provides labels only for its smaller  subsets. On the other hand, reinforcement learning provides `a kind of' supervision for the learning problem in terms of rewards or punishments to the algorithm. Due to their remote relevance to the tasks in Medical Imaging, we do not provide further discussion on these categories. Interested readers are directed to \cite{zhu2006semi} for semi-supervised learning, and to \cite{kaelbling1996reinforcement} for  reinforcement learning. 





\subsection{Standard Artificial Neural Networks}
\label{sec:ANN}
\begin{figure}[t]
\centering
\includegraphics[width=2.3in]{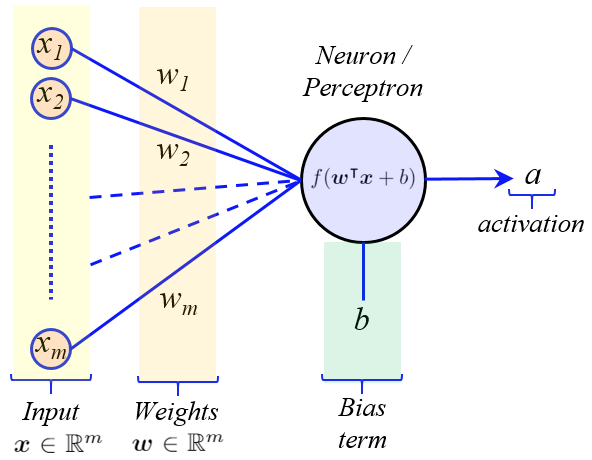}
\caption{Illustration of a single neuron/perceptron in a standard ANN. Each feature/coefficient `$x_i$ $\forall i \in \{1,...,m\}$' of the input signal  `$\boldsymbol{x}$' gets weighted by a corresponding weight `$w_i$'. A bias term `$b$' is added to the weighted sum `$\boldsymbol{ w}^{\intercal} \boldsymbol x$' and a non-linear/linear function $f(.)$ is applied to compute the activation `$a$'.  }
\label{fig:neuron}
\end{figure}

An Artificial Neural Network (ANN) is a hierarchical composition of basic computational elements known as \textit{neurons} (or   \textit{perceptrons}~\cite{rosenblatt1961principles}). Multiple neurons  exist at a single level of the hierarchy, forming a single \textit{layer} of the network. Using many layers in an ANN makes it \textit{deep}, see Fig.~\ref{fig:ANN}. A neuron performs the following simple computation:
\begin{align}
{a} = f({\boldsymbol w}^{\intercal} {\boldsymbol x} +  b),
\label{eq:activation}
\end{align}
where ${\boldsymbol x} \in \mathbb R^m$ is the input signal, ${\boldsymbol w} \in \mathbb R^m$ contains the neuron's weights, and $b \in \mathbb R$ is a \textit{bias} term. The symbol $f(.)$ denotes an \textit{activation function}, and the computed  $a \in \mathbb R$ is the neuron's \textit{activation signal}  or simply its \textit{activation}. Generally, $f(.)$ is kept non-linear to allow an ANN to induce complex non-linear computational models. The classic choices for $f(.)$ are the well-known `sigmoid' and  `hyperbolic tangent' functions. We depict a single neuron/perceptron in Fig.~\ref{fig:neuron}.

A neural network must learn the weight and bias terms in Eq.~(\ref{eq:activation}). The strategy used  to learn these parameters (i.e. back-propagation~\cite{rumelhart1986learning})  requires $f(.)$ to be a  differentiable function of its inputs. In the modern Deep Learning era, Rectified Linear Unit (ReLU)~\cite{krizhevsky2012imagenet} is widely used for this function, especially for non-standard ANNs such as CNNs (see Section~\ref{sec:CNN}). ReLU  is defined as $a = \max (0, {\boldsymbol w}^{\intercal} {\boldsymbol x} +  b)$. For the details beyond the scope of this discussion, ReLU allows more efficient and generally more effective learning of complex models as compared to the classic sigmoid and hyperbolic tangent activation functions.

It is possible to compactly represent the weights associated with all the neurons in a single layer of an ANN as a  matrix $\boldsymbol W \in \mathbb R^{p \times m}$, where `$p$' is the total number of neurons in that layer. This allows us to compute the activations of all the neurons in the layer at once as follows: 
\begin{align}
    \boldsymbol{a} = f(\boldsymbol{W}\boldsymbol{x} + \boldsymbol{b}),
\end{align}
where $\boldsymbol{a} \in \mathbb R^p$ now stores the activation values of all the neurons in the layer under consideration. Noting the heirarchical nature of ANNs, it is easy to see that the functional form of a  model induced by an $L$-layer  network can be given as:
\begin{align}
    \mathcal M(\boldsymbol{x}, \boldsymbol{\Theta}) = f_L( \boldsymbol{W}_L f_{L-1} ( \boldsymbol{W}_{L-1} f_{L-2}...(\boldsymbol{W}_1 \boldsymbol{x} + \\ \nonumber \boldsymbol{b}_1)+... +  \boldsymbol{b}_{L-1}) + \boldsymbol{b}_L),
\end{align}
where the subscripts denote the layer numbers. We collectively denote the parameters $\boldsymbol{W}_i, \boldsymbol{b}_i, \forall i\in\{1,...,L\}$ as $\boldsymbol\Theta$.

A neural network model can be composed by using different number of layers, having different number of neurons in each layer, and even having different activation functions for different layers. Combined, these choices determine the \textit{architecture}  of a neural network. The design variables of the architecture and those of the  learning algorithm are termed as \textit{hyper-parameters} of the network. Whereas the model parameters (i.e.~$\boldsymbol\Theta$) are learned automatically, finding the most suitable values of the hyper-parameter is usually a manual iterative process.
Standard ANNs are also commonly known as Multi-Layer Perceptrons (MLPs), as their layers are generally composed of standard neurons/perceptrons. One notable exception to this layer composition is  encountered at the very last layer, i.e.~\textit{softmax} layer used in  classification. In contrast to `independent' activation  computation by each neuron in a standard perceptron layer, softmax neurons compute activations that are normalized across all the activation values of that layer. Mathematically, the $i^{\text{th}}$ neuron of a softmax layer computes the activation value as:
\begin{align}
    a_i = \frac{e^{\boldsymbol{w}_i^{\intercal} \boldsymbol{a}_{L-1} + b_i}}{\sum\limits_{j=1}^{p} e^{\boldsymbol{w}_j^{\intercal} \boldsymbol{a}_{L-1} + b_j}}.
\end{align}

The benefit of normalizing the activation signal is that the output of the softmax layer can be interpreted as a \textit{probability} vector that encodes the \textit{confidence} of the network that a given sample belongs to a particular class. This interpretation of softmax layer outputs is a widely used concept in the related literature.

\subsection{Convolutional Neural Networks}
\label{sec:CNN}
In the context of DL techniques for image analysis, Convolution Neural Networks (CNNs)~\cite{krizhevsky2012imagenet}, \cite{lecun1998gradient}  are of the primary importance.  Similar to the standard ANNs, CNNs consist of multiple layers. However, instead of simple perceptron layers, we encounter three different kinds of layers in these networks (a)~Convolutional layers, (b)~Pooling layers, and (c)~Fully connected layers (often termed fc-layers). We describe these layers below, focusing mainly on the Convolutional layers that are the main source of strength for CNNs.

\paragraph{Convolutional layers}
The aim of convolution layers is to  learn weights of the so-called\footnote{Strictly speaking, the kernels in CNNs compute  cross-correlations. However, they are always referred to as `convolutional' by convention. Our subsequent explanation of the `convolution' operation is in-line with the definition of this operation used in the context of CNNs.} convolutional \textit{kernels/filters} that can perform convolution operations on images. Traditional image analysis  has a long history of using such filters to highlight/extract different image features, e.g. Sobel filter for detecting edges in images~\cite{sobel19683x3}. However, before CNNs, these filters needed to be designed  by manually setting the weights of the kernel in a careful manner. The breakthrough that CNNs provided is in the automatic learning of these weights under the  neural network settings.

\begin{figure}[t]
\centering
\includegraphics[width=3in]{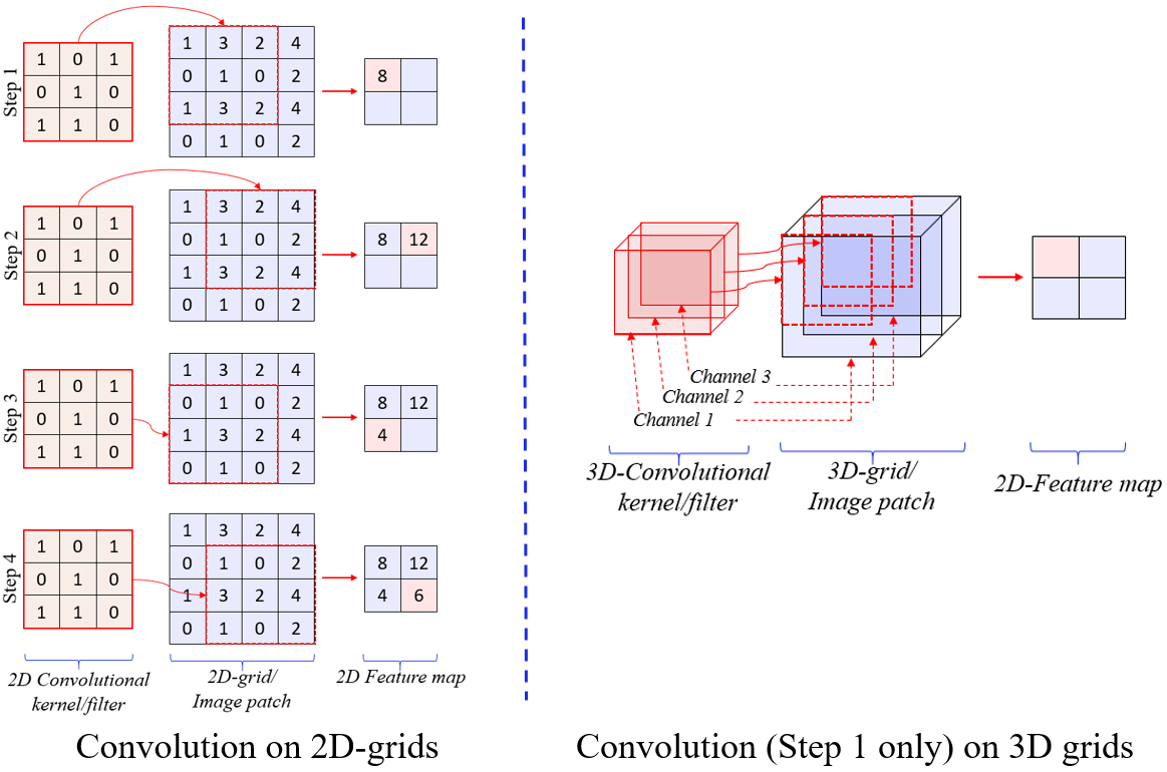}
\caption{Illustration of convolution operation for 2D-grids~(left) and 3D-volumes (right). Complete steps of moving window are shown for the 2D case whereas only step-1 is shown for 3D. In 3D, the convolved channels are combined by simple addition, still resulting in 2D feature maps. }
\label{fig:conv}
\end{figure}

We illustrate the convolution operation in Fig.~\ref{fig:conv}.
In 2D settings (e.g.~grey-scale images), this operation involves moving a small window (i.e.~kernel) over a 2D grid (i.e.~image). In each moving step, the corresponding elements of the two grids get multiplied and summed up to compute a scalar value. Concluding the operation results in another 2D-grid, referred to as the  \textit{feature/activation} map in the CNN literature. 
In 3D settings, the same steps are performed for  the individual pairs of the corresponding channels of the 3D volumes, and the resulting feature maps are simply added to compute a 2D map as the final output. Since color images have multiple channels,  convolutions in 3D settings are more relevant for the modern CNNs. However, for the sake of better understanding, we often discuss the relevant concepts using the 2D grids. These concept are readily transferable to the 3D volumes. 

\begin{figure}[t]
\centering
\includegraphics[width=3.5in]{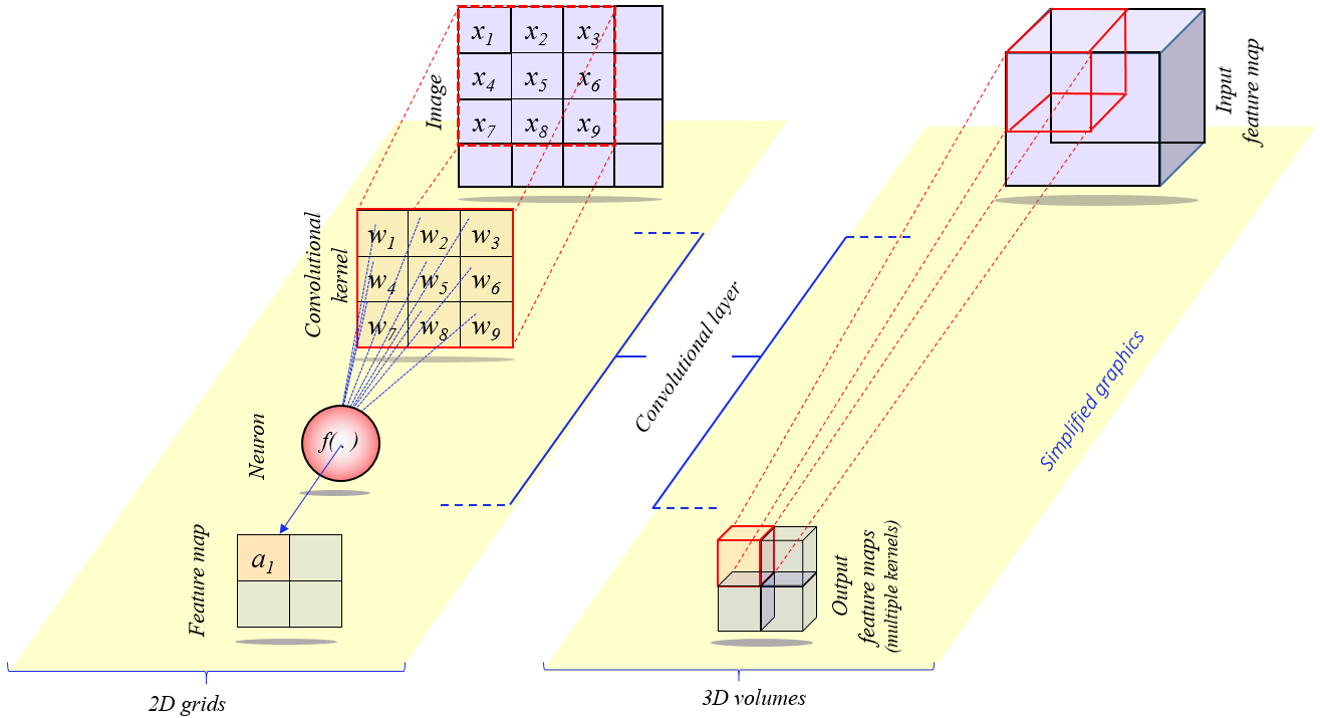}
\caption{Working of a convolutional layer. CNNs force kernel weights to become network parameters. (Left) In 2D grids, a single kernel moves over the image/input signal. (Right) A volume of multiple kernel moves over the input volume to result in an output volumes. }
\label{fig:convlayer}
\end{figure}

A convolutional layer of CNN forces the elements of kernels to become the network weights, see Fig.~\ref{fig:convlayer} for illustration.
In the figure, we can directly compute the activation $a_1$ (2D grids) using Eq.~(\ref{eq:activation}), where $\boldsymbol{x}, \boldsymbol{w} \in \mathbb R^9$ are vectors formed by arranging $w_1,..,w_9$ and $x_1,...,x_9$ in the figure. The figure does not show the bias term, which is generally ignored in the convolutional layers. It is easy to see that under this setting, we can make use of the same tools to learn the convolutional kernel weights that we use to learn the weights of a standard ANN. The same concept applies to the 3D volumes, with a difference that we must use `multiple' kernels to get a volume (instead of a 2D grid) at the output. Each feature map resulting from a kernel then acts as a separate channel of the output volume of a convolutional layer. It is a common practice in CNN literature to simplify the illustrations in 3D settings by only showing the input and output volumes for different layers, as we do in Fig.~\ref{fig:convlayer}.

From the perspective presented above, a convolutional layer may look very similar to a standard perceptron layer, discussed in Section~\ref{sec:ANN}. However, there are two major differences between the two. (1) Every input feature gets connected to its activation signal through the same kernel (i.e.~weights). This implies that all input features \textit{share} the kernel weights - called  \textit{parameter  sharing}. Consequently, the kernels try to adjust their weights such that they resonate well to the basic building patterns of the whole input signal, e.g.~edges for an image. 
(2)~Since the same kernel connects all input features to output features/activation, convolutional layers have very few parameters to learn in the form of kernel weights. This \textit{sparsity of connections} allows very efficient learning despite the high dimensionality of data - a feat not possible with standard \textit{densely} connected perceptron layers.

\paragraph{Pooling layers}
The main objective of a pooling layer is to reduce the width and height of the activation maps in CNNs.
The basic concept is to compute a single output value `$v$' for a small $n_p \times n_p$ grid in the activation map, where `$v$' is simply the maximum or average value of that grid in the activation map. Based on the used operation, this layer is often referred as \textit{max-pooling} or \textit{average-pooling} layer. Interestingly, there are no learnable parameters associated with a pooling layer. Hence, this layer is sometimes seen as a part of Convolutional layer. For instance, the popular VGG-16 network~\cite{simonyan2014very} does not see pooling layer as a separate layer, hence the name VGG-\textit{16}. On the other hand, other works, e.g.~\cite{sun2018deep} that use VGG-16 often count more than 16 layers in this network by treating the pooling layer as a regular network layer.   

\paragraph{Fully connected layers}
These layers are the same as the perceptron layers encountered in the standard ANNs.
The use of multiple Convolutional and Pooling layers in CNNs gradually reduces the size of resulting activation maps. Finally, the activation maps from a deeper layer are  re-arranged into a vector which is then fed to the fully connected (fc) layers. It is a common knowledge now that the activation vectors of fc-layers often serve as very good compact representations of the input signals (e.g.~images). 

Other than the above mentioned three layers, \textit{Batch Normalization} (BN)~\cite{ioffe2015batch} is another layer that is now a days  encountered more often in CNNs than in the standard  ANNs.   The main objective of this layer (with learnable parameters) is to control the mean and variance of the activation values of different network layers such that the induction of the overall model becomes more efficient. This idea is inspired by a long known fact that induction of ANN models generally becomes easier if the inputs are normalized to have zero mean and unit variance. The BN layer essentially applies a similar principle to the activations of deep neural networks.




\subsection{Recurrent Neural  Networks}
\begin{figure}[t]
\centering
\includegraphics[width=3in]{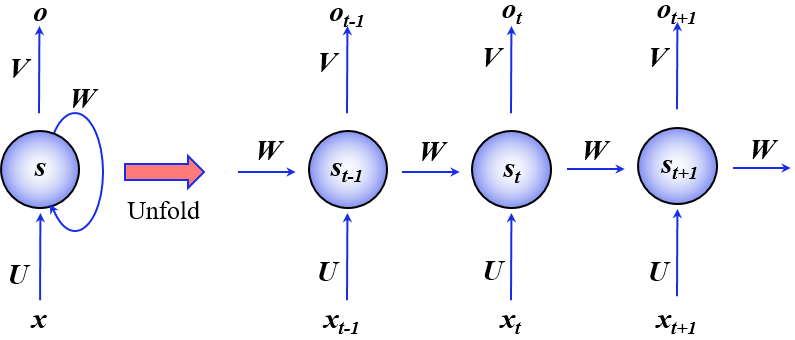}
\caption{Illustration of unfolded RNN~\cite{lecun2015deep}. A state $\boldsymbol{s}$ is shown with a single neuron for simplicity. At the ${t}^{\text{th}}$ time stamp, the network updates its state $\boldsymbol{s}_t$ based on the input $\boldsymbol{x}_t$ and the previous state. It optionally outputs a signal $\boldsymbol{o}_t$. $\boldsymbol{U}$, $\boldsymbol{V}$ and $\boldsymbol{W}$ are network parameters.    }
\label{fig:RNN}
\end{figure}

Standard neural networks assume that input signals  are independent of each other. However, often this is not the case. For instance, a word appearing in a sentence generally  depends on sequence of the words preceding it. Recurrent Neural Networks (RNNs) are designed to model such sequences.  
An RNN can be thought to maintain a `memory' of the sequence with the help of its internal states.
In Fig.~\ref{fig:RNN}, we show a typical RNN that is \textit{unfolded} - complete network is shown for the sequence. If the RNN has three layers,  it can model e.g.~ sentences that are three words long. In the figure, $\boldsymbol {x}_t$ is the input at the $t^{\text{th}}$ time stamp. For instance, $\boldsymbol{x}_t$ can be some quantitative representation of the $t^{\text{th}}$ word in a sentence. The memory of the network is maintained by the state $\boldsymbol{s}_t$ that is computed as:
\begin{align}
    \boldsymbol{s}_t = f(\boldsymbol{U}\boldsymbol{x}_t + \boldsymbol{W}\boldsymbol{s}_{t-1}),
\end{align}
where $f(.)$ is typically a non-linear activation function, e.g.~ReLU. The output at a given time stamp $\boldsymbol{o}_t$ is a function of a weighted version the network state at that time. For instance, predicting probability of the next word in a sentence can assume the output form $\boldsymbol{o}_t = \text{softmax}(\boldsymbol{V}\boldsymbol{s}_t)$, where `softmax' is the same operation discussed in Section~\ref{sec:ANN}.

One aspect to notice in the above equations is that we use the same weight matrices $\boldsymbol{U}, \boldsymbol{V}, \boldsymbol{W}$ at all time stamps. Thus, we are recursively performing the same operations  over an input sequence at multiple time stamps. This fact also inspires the name `Recursive' NN. It also has a significant implication that for an RNN we need a special kind of back-propagation algorithm, known as  \textit{back-propagation through time} (BTT). As compared to the regular back-propagation, BTT must propagate error recursively back to the previous time stamps. This becomes problematic for long sequences that involve too many time stamps. A phenomenon known as \textit{vanishing/exploding gradient} is the root cause of this problem. This has lead RNN researcher to focus on designing  networks that can handle longer sequences. Long Short-Term Memory (LSTM) network~\cite{hochreiter1997long} is currently a popular type of RNN that is found be reasonably effective for dealing with long sequences.   

LSTM networks have the same fundamental architecture of an RNN, however their hidden states are computed differently. The hidden units are commonly known as \textit{cells} in the context of LSTMs. Informally, a cell takes in the previous state and the input at a given time stamp and decides on what to remember and what to erase from its memory. The previous state, current input and the memory is then combined for the next time stamp.

\subsection{Using Neural Networks for Unsupervised Learning}
\label{sec:Unsupervised}
Whereas we assumed availability of the label for each data sample while discussing the basic concepts of neural networks in the preceding subsections, those concepts can also be  applied readily to construct neural networks to model data without labels. Here, we briefly discuss the mainstream frameworks that allow  to do so. It is noteworthy that this article  does not present neural networks as `supervised vs unsupervised' intentionally. This is because the core concepts of neural networks are generally better understood in   supervised settings. Unsupervised use of neural networks simply requires to employ the same ideas under different overall frameworks.

\subsubsection{Autoencoders}
\label{sec:AE}
The main idea behind autoencoders is to map an input signal (e.g.~image, feature vector) to itself using a neural network. In this process, we aim to learn a latent representation of the data that is more powerful for a particular task than the raw data itself. For instance, the learned  representation could cluster better than the original data. 
Theoretically, it is possible to use any kind of network layers in autoencoders that are used for supervised neural networks. The uniqueness of autoencoders comes in the output layer where the signal is the same as the input signal to the network instead of e.g.~a label vector in classification task.

Mapping a signal to itself can result in trivial models (learning identity mapping). Several techniques have been adopted in the literature to preclude this possibility, leading to different kinds of autoencoders. For instance, \textit{undercomplete} autoencoders ensure that the dimensionality of the latent representation is much smaller than the data dimension. In MLP settings, this can be done by using a small number (as compared to the input signal's dimension) of neurons in a hidden layer of the network, and use the activations of that layer as the latent representation. \textit{Regularized} autoencoders also impose sparsity on neuron connections~\cite{poultney2007efficient} and reconstruction of the original signal from its noisy vesion~\cite{vincent2010stacked} to ensure learning of useful latent representation instead of identity mapping. Variational autoencoders~\cite{kingma2013auto} and contractive autoencoders~\cite{rifai2011contractive} are also among the other popular types of autoencoders.  

\subsubsection{Generative Adversarial Networks} 
\label{sec:GANs}
Recent years have seen an extensive use of Generative Adversarial Networks (GANs)~\cite{goodfellow2014generative} in natural image analysis. GANs can be considered a variation of autoencoders that aim at mimicking the distribution generating the data. GANs are composed of two parts that are neural networks. The first part, termed  \textit{generator}, has the ability to generate a sample whereas the other, called \textit{discriminator} can classify the sample as a real or fake. Here, a `real' sample means that it is actually coming from the training data. The two networks essentially play a game where the generator tries to fool the discriminator by generating more and more realistic samples. In the process the generator keeps updating its parameters to produce better samples. The adversarial objective of the generator to fool the discriminator also inspires the name of GANs. In natural image analysis, GANs have been successfully applied for many tasks, e.g.~inducing realism in synthetic images~\cite{shrivastava2017learning}, domain adaption~\cite{bousmalis2017unsupervised}  and data completion~\cite{yeh2017semantic}. Such successful applications of GANs to image processing tasks also open new directions for medical image analysis tasks.


\subsection{Best practices in using CNNs for image analysis}
\label{sec:PopCNN}
 Convolutional Neural Networks (CNNs) form the backbone of the recent breakthroughs in image analysis. 
 To solve different problems in this area, CNN based models are normally used in three  different ways. (1) A network architecture is chosen and trained \textit{from scratch} using the available training dataset in an \textit{end-to-end} manner. (2) A CNN model \textit{pre-trained} on some large-scale dataset is \textit{fine-tuned} by further training the model for a few epochs using the data available for the problem at hands. This approach is more suitable when  limited training data is available for the problem under consideration. It is often termed  \textit{transfer learning} in the literature. (3) Use a model as a \textit{feature extractor} for the available images. In this case, training/testing images are passed through the network and the activations of a specific layer (or a combination of layers) are considered as image features. Further analysis is performed using those features. 
 
Computer Vision literature provides extensive studies to reflect on the best practices of exploiting CNNs in any of the aforementioned three manners. We can summarize the crux of these practices as follows. One should only consider training a model from scratch if the available training data size is very large, e.g.~ 50K image or more. If this is not the case, use transfer learning. If the training data is even smaller, e.g.~few hundred images, it may be better to use CNN only as a feature extractor. No matter which approach is adopted, it is better that the underlying CNN is inspired by a model that has already proven its effectiveness for a similar task. This is especially true for the `training from scratch' approach. We refer to the most successful recent CNN models in the Computer Vision literature in the paragraphs to follow. For transfer learning, it is better to use a model that is pre-trained on data/problem that is as similar as possible to the data/problem at hands. In the case of using CNN as a feature extractor, one should prefer a network with more representation power. Normally, deeper networks that are trained on very large datasets have this property. Due to their discriminative abilities, features extracted from such models are especially useful for classification tasks.

Starting from AlexNet in 2012~\cite{krizhevsky2012imagenet}, many complex CNN models have been developed in the last seven years. Whereas still useful, AlexNet is no longer considered a state-of-the-art network. A network still applied frequently is VGG-16~\cite{simonyan2014very} that was proposed in 2014 by the  Visual Geometry Group (VGG) of Oxford university. A later version of VGG-16 is VGG-19 that uses 19 instead of 16 layers of the  learnable parameters. Normally, the representation power of both versions are considered similar. Another popular network is GoogLeNet~\cite{szegedy2015going} that is also commonly known as `Inception' network. This network uses a unique type of layer called inception layer/block from which it drives its main strength. To date, four different versions of Inception~\cite{szegedy2016rethinking}, \cite{szegedy2017inception} have been introduced by the original authors, with each subsequent version having slightly better representation power (under a certain perspective) than its predecessor. ResNet~\cite{he2016deep} is another popular network that enables deep learning with models having more than hundred layers. It is based on a concept known as `residual learning', which is currently highly favored by Pattern Recognition community because it enables very deep networks. DenseNet~\cite{huang2017densely} also exploits the insights of residual learning to achieve the representation power similar to ResNet, but with a more compact network.

Whereas the above-mentioned CNNs are mainly  trained for image classification tasks, Fully Convolutional Networks (FCNs)~\cite{long2015fully} and U-Net~\cite{ronneberger2015u} are among the most popular networks for the task of image segmentation. Analyzing the architectures and hyperparamter settings of these networks can often reveal useful insights for developing new networks. In fact, some of these networks (e.g.~Inception-v4/ResNet~\cite{szegedy2017inception}) already rely on the insights from others (e.g.~ResNet~\cite{he2016deep}). The same practice can yield popular networks in the future as well. We draw further on the best practices of using CNNs for image analysis in  Section~\ref{sec:FD}.




\subsection{Deep Learning Programming Frameworks}
\label{sec:DLPF}
The rise of Deep Learning has been partially enabled by the public access to programming frameworks that implement the core techniques in this area in high-level programming languages. Currently, many of these frameworks are being continuously maintained by software developers and most of the new findings are incorporated in them rapidly. Whereas availability of appropriate Graphical Processing Unit (GPU) is desired to fully exploit these modern frameworks, CPU support is also provided with most of them to train and test small models. The frameworks allow their users to directly test different network architectures and their hyperparameter settings  etc.~without the need of actually implementing the operations performed by the layers and the algorithms that train them. The layers and related algorithms come pre-implemented in the libraries of the frameworks.

Below we list the popular Deep Learning frameworks in use now a days. We order them based on their current popularity in Computer Vision and Pattern Recognition community for the problems of image analysis, starting from the most popular. 
\begin{itemize}
    \item \href{https://www.tensorflow.org/}{\textit{Tensorflow}}~\cite{abadi2016tensorflow} is originally developed by Google Brain, it is fast becoming the most popular deep learning framework due to its continuous development. It provides Python and C++ interface.
    \item \href{https://pytorch.org/}{\textit{PyTorch}}~\cite{paszke2017automatic} is a Python based library supported by Facebook's AI Research. It is currently receiving significant attention due to its ability to implement dynamic graphs. 
    \item \href{https://caffe2.ai/}{\textit{Caffe2}}~\cite{jia2014caffe} builds on Caffe (see below) and provides C++ and Python interface. 
    \item \href{https://keras.io/}{\textit{Keras}}~\cite{chollet2015keras} can be seen as a high level programming interface that can run on top of Tensorflow and Theano~\cite{2016arXiv160502688short}. Although not as felxible as other frameworks, Keras is particularly popular for quickly developing and testing networks  using common network layers and algorithms. It is often seen as a gateway to deep learning for new users. 
    \item \href{http://www.vlfeat.org/matconvnet/}{\textit{MatConvNet}}~\cite{vedaldi2015matconvnet} is the most commonly used public deep learning library for Matlab. 
    \item \href{https://caffe2.ai/}{\textit{Caffe}}~\cite{jia2014caffe} was originally developed by UC Berekely, providing C++ and Python interface. Whereas Caffe2 is now fast replacing Caffe, this framework is still in use because public implementations of many popular networks are available in Caffe. 
    \item \href{http://deeplearning.net/software/theano/}{\textit{Theano}}~\cite{2016arXiv160502688short} is a  library of Python to implement deep learning techniques that is developed and supported by MILA, University of Montreal.   
    \item \href{https://github.com/torch/torch7}{\textit{Torch}}~\cite{collobert2002torch} is a library and scripting language based on Lua. Due to its first release in 2011, many first generation deep learning models were implemented using Torch. 
\end{itemize}

The above is not an exhaustive list of the frameworks for Deep Learning. However, it covers those frameworks that are currently widely used in  image analysis. It should be noted, whereas we order the above list in terms of the `current' trend in popularity of the frameworks, public implementations of many networks proposed in 2012 - 2015 are originally found in e.g.~ Torch, Theano or Caffe. This also makes those frameworks equally important. However, it is often possible to find public implementations of legacy networks in e.g.~Tensorflow, albiet not by the original authors.

\section{Deep Learning Methods In Medical Image Analysis}
\label{sec:DLMethods}
\begin{figure*}[t]
\centering
\includegraphics[width=\textwidth]{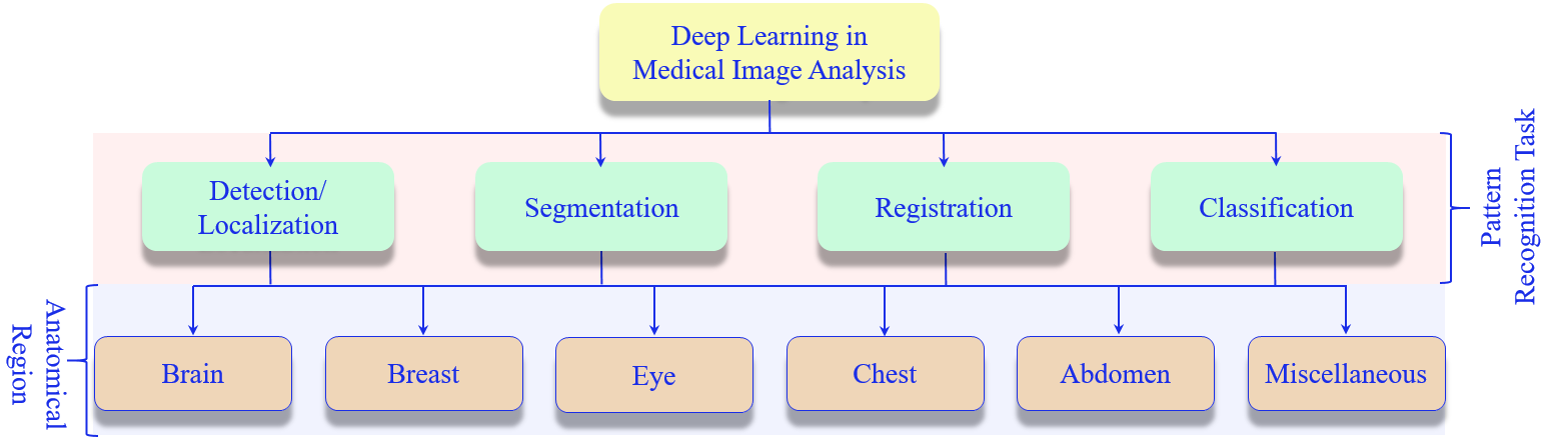}
\caption{Taxonomy of literature review: The contributions exploiting Deep Learning technology are first categorized  according to the underlying Pattern Recognition tasks. Each category is than sub-divided based on the human anatomical region studied in the papers. }
\label{fig:Tax}
\end{figure*}

In this Section, we review the recent contributions in Medical Image Analysis that exploit the Deep Learning technology. 
We mainly focus on the research papers published after December 2017, while briefly mentioning the more influential contributions from the earlier years. For a comprehensive review of the literature earlier than the year 2018, we collectively  recommend the following articles~\cite{sahiner2018deep, litjens2017survey, ker2018deep}. Taking more of a Computer Vision/Machine Learning perspective, we first categorize  the existing literature under  `Pattern Recognition' tasks. The literature pertaining to each task is then further sub-categorized based on the human anatomical regions. The taxonomy of our literature review is depicted in Fig.~\ref{fig:Tax}.

\subsection{Detection/Localization}
\label{sec:Detect}
The main aim of \textit{detection} is to identify a particular region of interest in an image an draw a bounding box around it, e.g.~brain tumor in MRI scans. Hence, \textit{localization} is also another term used for the detection task. In Medical Image Analysis, detection is more commonly referred to as Computer Aided Detection (CAD). CAD systems are aimed at detecting  the earliest signs of abnormality in patients. 
Lung and breast cancer detection can be considered as the common applications of CAD. 

\subsubsection{Brain}
\label{sec:S_brain}
For the anatomic region of brain, Jyoti \textit{et al.}~\cite{islam2018early} employed a CNN for the detection of Alzheimer's Disease (AD) using the MRI images of OASIS data set~\cite{marcus2010open}.
The authors built on two baseline CNN networks, namely Inception-v4~\cite{szegedy2017inception} and ResNet~\cite{he2016deep}, to categorize four classes of AD. These classes include moderate, mild, very mild and  non-demented patients. The accuracies reported by the authors for these classes are $33\%$, $62\%$, $75\%$, and $99\%$, respectively. 
It is claimed that the proposed method does not only perform well on the used dataset, but it also has the potential to generalize to ADNI dataset~\cite{petersen2010alzheimer}. 
Chen~\textit{ et al.}~\cite{chen2018unsupervised} proposed an unsupervised learning approach using an  Auto-Encoders (AE). The authors investigated lesion  detection using Variational Auto Encoder (VAE)~\cite{kingma2013auto} and Adversarial Auto Encoder (AAE)~\cite{makhzani2015adversarial}. The analysis is carried out on BRATS 2015 datasets, demonstrating good results for the Aera Under Curve (AUC) metric. 

Alaverdyan \textit{et al.}~\cite{alaverdyan2018regularized} used a deep neural network for epilepsy lesion detection in multiparametric MRI images. They also stacked convolution layers in an auto-encoders fashion and trained their network using the patches of the original images. Their model was trained using the data from 75 healthy subjects in an unsupervised manner.
For the automated brain tumor detection in MR images 
Panda\textit{et al.}~\cite{panda2019automated} used  discriminative clustering method to segregate the vital regions of brain such as Cerebro Spinal Fluid (CSF), White Matter (WM) and Gray Matter (GM). In another study of automatic detection in MR images ~\cite{laukamp2019fully}, Laukamp\textit{et al.} used multi-parametric deep learning model for the detection of meningiomas in brain. 

\subsubsection{Breast}
\label{sec:S_breast}
In assessing cancer spread, histopathological analysis of
Sentinel Lymph Nodes (SLNs) becomes important for the task of cancer staging. Bejnordi \textit{et al.}~\cite{bejnordi2017diagnostic} analyzed deep learning techniques for metastases detection in eosin-stained tissues and hematoxylin tissue sections of lymph nodes of the subjects with  cancer. The computational results are  compared with human pathologist diagnoses. Interestingly, out of the 32 techniques analysed, the top 5 deep learning algorithms arguably out-performed eleven pathologists.  

Chiang \textit{et al.}~\cite{chiang2019tumor} developed a CAD technique based on a 3D CNN for breast cancer detection using Automated whole Breast Ultrasound (ABUS) imaging modality. In their approach, they first extracted Volumes of Interest (VOIs) through a sliding window technique, then the 3D CNN was applied and tumor candidates were selected based on the probability resulting from the application of 3D CNN to VOIs.  In the experiments  171 tumors are used for testing, achieving  sensitivities of up to $95\%$.
Dalmics \textit{et al.}~\cite{dalmics2018fully} proposed a CNN based CAD system to detect breast cancer in MRI images. They used 365 MRI scans for training and testing, out of which 161 were malignant lesions. They claimed the achieved sensitivity obtained by their technique to be better than the existing  CAD systems.
For the detection of breast mass in mammography images, Zhang \textit{et al.}~\cite{zhang2018breast} developed a Fully Convolutional Network (FCN) based end-to-end heatmap regression technique. They  demonstrated that mammography data could be used for digital breast tomosynthesis (DBT) to improve the detection model. They used transfer learning by fine tunning an FCN model on  mammography images. The approach is tested on  tomosynthesis data with  40 subjects, demonstrating better performance as compared to the model trained from  scratch on the same data. 

\subsubsection{Eye}
\label{sec:S_eye}
For the anatomical region of eye, Li \textit{et al.}~\cite{li2019fully} recently employed a deep transfer learning approach which fine tunes the  VGG-16 model~\cite{simonyan2014very} that is pretrained on ImageNet~\cite{deng2009imagenet} dataset. To detect and classify Age-related Macular Degeneration (AMD) and Diabetic Macular Edema (DME) diseases in eye, they used 207,130 retinal Optical Coherence Tomography (OCT) images. The proposed method achieved $98.6\%$ prediction detection accuracy in retinal images with $100\%$. 
Ambramoff \textit{et al.}~\cite{abramoff2016improved} used a CNN based technique to detect Diabetic Retinopathy (DR) in fundus images. They assessed the device IDx-DR X 2.1 in their study using a public dataset~\cite{gargeya2017automated} and achieve an AUC score of $0.98$. Schlegl \textit{et al.}~\cite{schlegl2018fully}  employed deep learning for the detection and quantification of Intraretinal Cystoid Fluid (IRC) and Subretinal Fluid (SRF) in retinal images. They employed an auto encoder-decoder formation of CNNs, and used 1,200 OCT retinal images for the  experiments,  achieving AUC of $0.92$ for SRF and AUC of $0.94$ for IRC.

Deep learning is also being increasingly used for diagnosing retinal diseases~\cite{kermany2018identifying},\cite{us2018fda}. Li \textit{ et al.}~\cite{li2018efficacy} trained a deep learning model based on the Inception architecture~\cite{szegedy2015going} for the identification of Glaucomatous Optic Neuropathy (GON) in retinal images. Their model achieved AUC of $0.986$ for  distinguishing healthy from GON eyes. Recently, Christopher \textit{et al.} ~\cite{christopher2018performance} also used transfer learning with VGG16, Inception v3, and ResNet50 models for the identification of GON. They used pre-trained  models of ImageNet. For their experiments, they used 14,822 Optic Nerve Head (ONH) fundus images of GON or healthy eyes. The achieved best performance for identifying moderate-to-severe GON in the eyes was reported to be AUC value $0.97$ with $90\%$ sensitivity and $93\%$ specificity. Khojasteh \textit{et al.}~\cite{khojasteh2019exudate} used pre-trained ResNet-50 on  DIARETDB1~\cite{kalviainen2007diaretdb1} and e-Ophtha~\cite{decenciere2013teleophta} datasets for the detection of excudates in the  retinal images. They reported an accuracy of $98\%$ with $99\%$ sensitivity  of detection on the used data.

\subsubsection{Chest}
\label{sec:S_chest}

For the pulmonary nodule detection in lungs in Computed Tomography (CT) images, Zhu \textit{et al.}~\cite{zhu2018deepem} proposed a deep network called DeepEM. This network uses a 3D CNN architecture that is  augmented with an Expectation-Maximization (EM) technique  for the noisily labeled data of Electronic Medical Records (EMRs).
They used the EM technique to train their model in an  end-to-end manner.
Three datasets were  used in their study, including;  the LUNA16 dataset~\cite{setio2017validation} - the largest publicly available dataset for supervised nodule detection, NCI NLST dataset\footnote{\url{https://biometry.nci.nih.gov/cdas/datasets/nlst/}} for weakly supervised detection and Tianchi Lung Nodule Detection dataset.

For the detection of artefacts in Cardiac Magnetic Resonance (CMR) imaging, Oksuz \textit{et al.}~\cite{oksuz2018deep}  also proposed a CNN based technique. Before training the model, they performed  image pre-processing by normalization and region of interest (ROI) extraction. The authors used a CNN architecture with 6-convolutional layers (ReLU activations) followed by 4-pooling layers, 2 fc layers and a softmax layer to estimate the motion artefact labels. They showed good performance for the classification of motion artefacts in videos. The authors essentially  built on the insights of~\cite{tran2015learning} in which video classification is  done using a spatio-temporal 3D CNN.

Zhe \textit{et al.}~\cite{li2017thoracic} proposed a technique for the localization and identification of thoraic diseases in public database NIH X-ray\footnote{\url{https://www.kaggle.com/nih-chest-xrays/data}} that comprises 120 frontal view X-ray images with 14 labels. Their model performs the tasks of localization and identification simultaneously. They used the popular ResNet~\cite{he2016deep} architecture to build the computational model. In their model, an input image is passed through the CNN for feature map extraction, then a max pooling or bi-linear interpolation layer is used for resizing the input image by a patch slicing layer.  Afterwards, fully convolutional layers are used to eventually perform the recognition. For training, the authors exploit the framework of Multi-Instance Learning (MIL), and in the testing phase, the model predicts both labels and class specific localization details. 
Yi \textit{et al.}~\cite{yi2018automatic} presented a scale recurrent network for the detection of catheter in X-ray images. Their network architecture is  organised in an auto encoder-decoder manner. In another study~\cite{masood2018computer}, Masood \textit{et al.} proposed a deep network, termed DFCNet, for the automatic computer aided lung pulmonary detection.

Gonzalez \textit{et al.}~\cite{gonzalez2018disease} proposed a deep network for the detection of Chronic Obstructive Pulmonary Disease (COPD) and Acute Respiratory Disease (ARD) prediction in CT images of smokers. They trained a CNN using 7,983 COPDGene cases and used logistic regression for COPD detection and ARD prediction.
In another study~\cite{gonzalez2018improving}, the same group of researchers used deep learning for weakly supervised lession localization.
Recently, Marsiya \textit{et al.} \cite{winkels20183d} used NLST and  LDIC/IDRI~\cite{armato2011lung}  datasets for  lung nodule detection in CT images. They proposed a 3D Group-equivariant Convolutional Neural Network (G-CNN) technique for that purpose. The proposed method was exploited for fast positive reduction in pulmonary lung nodule detection. The authors claim their method performs on-par with standard CNNs while trained using ten times less data.

\subsubsection{Abdomen}
\label{sec:S_abdomen}


Alensary \textit{et al.}~\cite{alansary2018evaluating} proposed a deep reinforcement learning technique for the detection of multiple landmarks with ROIs in 3D fetal head scans. 
Ferlaino \textit{et al.}~\cite{ferlaino2018towards} worked on plancental histology using deep learning. They classified five different classes with an accuracy of $89$\%. Their model also learns deep embedding
encoding phenotypic knowledge that classifies five different 
cell populations and learns inter-class variances of phenotype. 
Ghesu \textit{et al.}~\cite{ghesu2019multi} used a large data of 1,487 3D CT scans for the detection of anatomic sites, exploiting multi-scale deep reinforcement learning. 

\begin{table*}[t]
\centering
\caption{Summary of highly influential papers appeared in 2016 and 2017 (based on Google Scholar citation index in January 2019) that exploit deep learning for Detection/Localization in Medical Image Analysis.} 
\label{tab:detect}
\begin{tabular}{|c|c|c|c|c|c|}
\hline
\textbf{Reference}    &\textbf{Anatomic Site}
& \textbf{Image Modality}     & \textbf{Network type}  & \textbf{Data}&
\textbf{Citations}
\\
\hline \hline

Shin \textit{ et al.}~\cite{hoo2016deep} (2016)&  Lung        
&      CT
&       CNN             & - & 783         \\ \hline
 Sirinukunwattana \textit{ et al.}~\cite{sirinukunwattana2016locality} (2016) &         Abdomen          &  Histopathology                & CNN                &   20000+ images           & 252                     \\ \hline
Setio \textit{ et al.}~\cite{setio2016pulmonary} (2016)  & Lung     &      CT 
&    CNN & 888 images                  &     247    \\ \hline
Xu \textit{ et al.}~\cite{xu2016stacked} (2016) & Breast                  &   Histopathology                &  AE               &  500 images  &  220                   \\ \hline
Wang \textit{ et al.}~\cite{wang2016deep} (2016) & Breast                  &   Histopathology                &                CNN &  400 images  &  182                  \\ \hline
Kooi \textit{ et al.}~\cite{kooi2017large} (2017) & Breast                  &   Mammography                &  CNN              &  45000 images  &  162                   \\ \hline
Rajpurkar \textit{ et al.}~\cite{rajpurkar2017chexnet} (2017) & Chest                 &   X-ray                &  CNN               &  100000+ images  &  139                  \\ \hline
Liu \textit{ et al.}~\cite{liu2017detecting} (2017) & Breast                  &   Histopathology                & CNN         &  400 slides &  98                 \\ \hline

Ghafoorian \textit{ et al.}~\cite{ghafoorian2017deep} (2017)  & Brain
&     MRI             &   CNN                &     LUNA16 ISBI 2016        & 90                 \\ \hline
Dou \textit{ et al.}~\cite{dou2017multilevel} (2017)  & Lung
&     CT            &  3D CNN                &      1075 images        & 40                 \\ \hline
 Zhang \textit{ et al.}~\cite{zhang2017detecting} (2017)          & Brain
&   MRI                 & FCN              &     700 subjects    &  32              \\ \hline

\end{tabular}
\end{table*}

Katzmann \textit{et al.}~\cite{katzmann2018predicting} proposed a deep learning based technique for the estimation of Colorectal Cancer (CRC) in CT tumor images for early treatment. Their model achieved high accuracies in growth and survival prediction.
Meng \textit{et al.}~\cite{meng2018automatic} formulated an automatic shadow detection technique in 2D ultrasound images using weakly supervised annotations. Their method highlights the shadow regions which is particularly useful for the segmentation task. Horie \textit{et al.}~\cite{horie2019diagnostic} recently applied a CNN technique for easophagal cancer detection. They used 8,428 WGD images and attained $98\%$ results for the  sensitivity.
Yasaka \textit{et al.}~\cite{yasaka2017deep} used a deep CNN architecture for the diagnosis of three different phases (noncontrast-agent enhanced, arterial, and delayed) of masses of liver in dynamic CT images.

\subsubsection{Miscellaneous }
\label{sec:S_misc}
 Zhang \textit{et al.}~\cite{zhang2018accurate} achieved $98.51$\% accuracy and a localization error $2.45$mm for the detection of inner ear in CT images. They used $3$D U-Net~\cite{cciccek20163d} to map the whole 3D image which consists of multiple convolution-pooling layers that convert the raw input image into the low resolution and highly abstracted feature maps. They applied false positive suppression technique in the training process and used a shape based constraint during training. 
Rajpurkar \textit{et al.}~\cite{rajpurkar2017mura} recently released a data set MURA which consists of 40,561 images from 14,863 musculoskeletal  studies labeled by radiologists as either normal or abnormal. The authors used CNN with 169-layers for the detection of normality and abnormality in each image study. Li \textit{et al.}~\cite{li2018cancer} proposed a Neural Conditional Random Field (NCRF) technique for the metastasis cancer detection in whole slide images. Their model was trained end-to-end using back-propagation and it obtained successful FROC score of 0.8096 in testing using Camelyon16 dataset~\cite{bandi2018detection}. 
Codella \textit{et al.}~\cite{codella2018skin} recently organized a  challenge  at  the  International  Symposium  on  Biomedical  Imaging  (ISBI), 2017  for skin lession analysis for melanoma detection. The challenge task provided  2,000 training images, 150 validation images, and 600 images for testing. It eventually published the results of 46 submission. We refer to \cite{codella2018skin} for further details on the challenge itself and the submissions.
We also mention few techniques in Table~\ref{tab:detect} related to the task of detection/localization. These methods appeared in the literature in the years 2016-17. Based on  Google Scholar's citation index, these methods are among the highly influential techniques in the current related literature. This article also provides similar summaries of the highly influential papers from the years 2016-17 for each pattern recognition task considered in the Sections to follow.

\subsection{Segmentation}
\label{sec:seg}
In Medical Image Analysis, deep learning is being extensively used for image segmentation with different modalities, including Computed Tomography (CT), X-ray, Positron-Emission Tomography (PET), Ultrasound, Magnetic Resonance Imaging (MRI) and Optical Cohrence Tomography (OCT) etc. Segmentation is the process of partitioning an image into different meaningful segments (that share similar characteristics) through automatic or semi-automatic outlining of the boundaries within the image. In medical imaging, these segments usually commensurate to different tissue classes, pathologies, organs or some other biological structure~\cite{forouzanfar2010parameter}.
\subsubsection{Brain}
\label{sec:S_brain}
Related to the anatomical region of brain, Dey \textit{et al.}~\cite{dey2018compnet} trained a complementary segmentation network, termed CompNet, for skull stripping in MRI scans for normal and pathological brain images.  The OASIS dataset~\cite{marcus2007open} was used for the training purpose. In their approach, the features used for segmentation are learned using an   encoder-decoder network trained from the images of brain tissues and its complimentary part outside the brain.
The approach is compared with a plain U-Net and a dense U-Net~\cite{kleesiek2016deep}. The accuracy achieved by the CompNet for  the normal images is $98.27\%$, and for the pathological images is $97.62\%$. These results are better than those achieved by~\cite{kleesiek2016deep}.

Zaho \textit{et al.}~\cite{zhao2018deep}  proposed a deep learning technique for brain tumor segmentation by integrating Fully Convolutional Networks (FCNs) and Conditional Random Fields (CRFs) in a combined framework to achieve segmentation with appearance and spatial consistency. They trained 3 segmentation  models using 2D image patches and slices. First, the training is performed for the FCN using image patches, then CRF is trained with a Recurrent Neural Network (CRF-RNN) using image slices. During this phase, the  parameters of the FCN are fixed. Afterwards, FCN and CRF-RNN parameters are jointly fined tuned using the image slice. 
The authors used the MRI image data  provided by Multimodal Brain Tumor Image Segmentation Challenge (BRATS) 2013, BRATS 2015 and BRATS 2016. 
In their work, Nair \textit{et al.}~\cite{nair2018exploring} used a 3D CNN approach for the segmentation and detection of Multiple Sclerosis (MS) lesions in MRI sequences.
Roy \textit{et al.}~\cite{roy2018inherent} used voxel-wise Baysian FCN for the whole brain segmentation by using Monte Carlo sampling. They demonstrated high accuracies on four datasets, namely MALC, ADNI-29, CANDI-13, and IBSR-18.
Robinson \textit{ et al.}~\cite{robinson2018subject} also proposed a real-time deep learning approach for the cardiavor MR segmentation.

\subsubsection{Breast}
\label{sec:S_breast}
In their study, Singh \textit{et al.}~\cite{singh2018conditional} described a conditional Generative Adversarial Networks (cGAN) model for breast mass segmentation in mammography. Experiments were conducted on Digital Database for Screening Mammography (DDSM) public dataset and private dataset of mammograms from Hospital Universitari Sant Joan de Reus-Spai.
They additionally used a simpler CNN to classify the segmented tumor area into different shapes (irregular, lobular, oval and round) and achieved an accuracy of $72\%$ on DDSM dataset.
Zhang \textit{ et al.} ~\cite{zhang2018automatic} exploited deep learning for image intensity normalization in breast segmentation task. They used 460 subjects from Dynamic Contrast-enhanced Magnetic Resonance Imaging (DCEMRI). Each subject contained one T1 weighted pre-contrast and three T1 weighted post-contrast images.
Men \textit{et al.}~\cite{men2018fully} trained a deep dilated ResNet for segmentation of Clinical Target Volume (CTV) in breasts.
Lee \textit{et al.}~ \cite{lee2018automated} based on fully convolution neural network proposed an automated segmentation technique for the breast density estimation in mammograms. For the evaluation of their approach,  they used full-field digital screening mammograms of 604 subjects. They fine tuned the pre-trained network for breast density segmentation and estimation. The Percent Density (PD) estimation by their approach showed similarities with BI-RADS density assessment by radiologists and outperformed the then state-of-the-art computational approaches.

\subsubsection{Eye}
\label{sec:S_eye}
Retinal blood image segmentation is considered an important and a challenging task in retinopathology. Zhang\textit{ et al.}~\cite{zhang2018deep} used a deep neural network for this purpose by exploiting the U-Net architecture~\cite{ronneberger2015u} with residual connection. They shown their results on three public datasets STARE~\cite{staal2004ridge}, CHASEDB1~\cite{fraz2012ensemble} and DRIVE~\cite{hoover2000locating} and achieved an AUC value of $97.99$\% for the DRIVE dataset. De \textit{et al.}~\cite{de2018clinically} applied deep learning for the retinal tissue segmentation. They used 14,884 three dimensional OCT images for training their network. Their approach is claimed to be device independent - it maintains segmentation accuracy while using  different device data. In another study of retinal blood vessels, Jebaseeli \textit{et al.}~\cite{jebaseeli2019segmentation} proposed a method to enhance the quality of retinal vessel segmentation. They analysed the severity level of diabetic retinopathy. Their proposed method, Deep Learning Based SVM (DLBSVM) model uses DRIVE, STARE, REVIEW, HRF, and DRIONS databases for training.
Liu \textit{et al.}~\cite{liu2019semi} proposed a semi-supervised learning for retinal layer and fluid region segmentation in retinal OCT B-scans. Adversarial technique was exploited for the unlabeled data. Their technique resembles the U-Net fully convolutional architecture.

\subsubsection{Chest}
\label{sec:S_chest}

Duan \textit{et al.}~\cite{duan2018deep} proposed a Deep Nested Level Set (DNLS) technique for the muti-region segmentation of cardiac MR images in patients with Pulmonary Hypertension (PH). They compared their approach with a CNN method~\cite{bai2017human} and the conditional random field (CRF) CRF-CNN approach~\cite{krahenbuhl2011efficient}. DNLS is shown to outperform those  techniques for all anatomical structures, specifically for myocardium. However, it requires more computations than \cite{bai2017human} which is the fastest method among the tested approaches. 
Bai \textit{et al.}~\cite{bai2018recurrent} used FCN and RNN for the pixel-wise segmentation of Aortic sequences in MR images. They trained their model in an end-to-end manner from sparse annotations by using a weighted loss function. The proposed method consists of two parts, the first  extracts features from the FCN using a U-Net architecture~\cite{ronneberger2015u}. The second feeds these features to an RNN for segmentation. 
Among the used 500 Arotic MR images (provided by the UK Biobank), the study used random 400 images for training, and the rest were used for testing the models.
Another recent study on semi supervised myocardiac segmentation has been conducted by Chartsias \textit{et al.}~\cite{chartsias2018factorised}, which was presented as an oral paper in MICCAI 2018. Their proposed network, called Spatial Decomposition Network (SDNet), model 2D input images in two representations, namely spatial representation of myocardiac as a binary mask and a latent representation of the remaining features in a vector form. While not being a fully supervised techniques, their method still achieves  remarkable results for the segmentation task.

Kervadec \textit{et al.}~\cite{kervadec2018constrained} proposed a CNN based ENet~\cite{paszke2016enet} constrained loss function for segmentation of weakly supervised Cardiac images. They achieved 90\% accuracy on the public datasets of 2017 ACDC challenge\footnote{\url{https://www.creatis.insa-lyon.fr/Challenge/acdc/}}. Their approach is closes the gap between weakly and fully supervised segmentation in semantic medical imaging.  In another study of cariac CT and MRI images for multi-class image segmentation, Joyce \textit {et al.}~\cite{joyce2018deep} proposed an  adversarial approach consisting of a shallow UNet like method. They also  demonstrated improved segmentation with an unsupervised cost. 

LanLonde \textit{et al.}~ \cite{lalonde2018capsules} introduced a CNN based technique, termed SegCaps, by exploiting the capsule networks \cite{sabour2017dynamic} for object segmentation. The authors exploited the  LUNA16 subset of the LIDC-IDRI database and demonstrated the effectiveness of their method for analysing CT lung scans. It is shown that the method achieves better segmentation performance as compared to the popular U-Net.
The SegCaps are able to handle large images with size $512\times512$. In another study~\cite{nam2018lung} of lung cancer segmentation using the LUNA16 dataset, Nam \textit{et al.} proposed a CNN model using 24 convolution layers, 2 pooling, 2 deconvolutional layers and one fully connected layer. Similarly,  
Burlutskiy \textit{et al.}~\cite{burlutskiy2018deep} developed a deep learning framework for lung cancer segmentation. They trained their model using the scans of 712 patients and tested on the scans of 178 patients of fully annotated Tissue Micro-Arrays (TMAs). Their model is aimed at finding  high potential cancer areas in TMA cores. 

\subsubsection{Abdomen}
\label{sec:S_abdomen}
Roth \textit{et al.}~\cite{roth2018deep}  built a 3D FCN model for automatic semantic segmentation of 3D images. The model is trained on clinical Computed Tomography (CT) data, and it is shown to perform  automated multi-organ segmentation of abdominal CT with $90\%$ average Dice score across all targeted organs. 
A CNN method, termed Kid-Net, is  proposed for kidney vessels; artery, vein and collecting system (ureter) segmentation by Taha \textit{et al.}~\cite{taha2018kid}. 
Their model is trained in an end-to-end fashion using 3D CT-volume patches.
One promising claim made by the authors  is that their method reduces kidney vessels segmentation time from hours to minutes.
Their approach uses feature down-sampling and up-sampling to achieve higher classification and localization accuracies.
Their network training methodology also handles unbalanced data, and focuses on reducing false positives.
It is also claimed that the proposed method enables high-resolution segmentation with a limited memory budget. 
The authors exploit the findings in~\cite{milletari2016v} for that purpose.

Oktay \textit{ et al.}~\cite{oktay2018attention} recently presented an `attention gate'  model to automatically find the target anatomy of different shapes and sizes. They essentially extended the U-Net model to an attention U-Net model for pancreas segmentation. Their model can be  utilized for organ localization and detection tasks. They used 120 images of CT for training their model, and 30 images for testing. Overall, the algorithm achieves good performance with $2$ to $3$\% increase in dice score as compared to the existing methods.
A related research on pancreas segmentation had been conducted  previously using dense connection by Gibson \textit{et al.}~\cite{gibson2017towards} and sparse convolutions by Heinrich \textit{ et al.}~\cite{heinrich2018obelisk},   \cite{heinrich2017briefnet,heinrich2018ternarynet}. 
For multi-organ  segmentation (i.e lung, heart, liver, bone) in unlabeled X-ray images, Zhang \textit{et al.}~\cite{zhang2018task} proposed a Task Driven Generative Adversarial Network (TD-GAN) automated technique. This is an unsupervised end-to-end method for medical image segmentation. They fine tuned a dense image-to-image network (DI2I)~\cite{huang2017densely,zhu2017unpaired} on synthetic Digitally Reconstructed Radiographs (DRRs) and  X-ray images. 
In another study of multi organ segmentation, Tong \textit{et al.}~\cite{tong2018fully} proposed an FCN  with a shape representation model. Their experiments were carried out on H$\&$N datasets of volumetric CT scans. 

 Yang \textit{et al.}~\cite{yang2017automatic} used a conditional Generative Adversarial Network (cGAN) to segment the human liver in 3D CT images.
 Lessmann \textit{et al.}~ \cite{lessmann2018iterative} proposed an FCN based technique for the automatic vetebra segmentation in CT images. The underlying architecture of their network is inspired by U-Net. Their model is able to process a patch size of $128\times128\times128$ voxels. It achieves $95.8\%$ accuracy for classification and $92.1\%$ for segmentation in the spinal images used by the authors.
Jin \textit{et al.}~\cite{jin2018ct} proposed a 3D CGAN to learn lung nodules conditioned on a Volume Of Interest (VOI) with an erased central region in 3D CT images. They trained their model on 1,000 nodules taken from LIDC dataset.
The proposed CGAN was further used to generate a dataset for Progressive Holistically Nested Network (P-HNN) model~\cite{harrison2017progressive} which demonstrates improved segmentation performance.

\subsubsection{Miscellaneous }
\label{sec:S_misc}
For memory and computational efficiency,   Xu \textit{et al.}~\cite{xu2018quantization} applied a  quantization mechanism to FCNs for the  segmentation tasks in Medical Image Analysis. They also used quantization to mitigate the over fitting issue for better performance. The effectiveness of the developed method is demonstrated for   2015 MICCAI Gland Challenge dataset~\cite{sirinukunwattana2017gland}. 
As compared to~\cite{yang2017suggestive} their method improves the results by up to $1\%$ with $6.4\times$ reduction in the memory usage.
Recently, Zhao \textit{et al.}~\cite{zhao2018deep} proposed a deep learning technique for $3$D image instance segmentation. Their model is trainable with weak annotations that  needs 3D bounding boxes for all instances and full voxel annotations for only a small fractions of instances. 
Liu \textit{et al.}~\cite{liu2018deep} employed a novel deep reinforcement learning approach for the segmentation and classification of surgical gesture. Their approach performs well on JIGSAW dataset in terms of edit score as compared to previous similar works.
Arif \textit{et al.} ~\cite{al2018spnet} presented a deep FCN model called SPNet, as shape predictor for object segmentation. The X-ray images used in their study are of cervical vertebra. 
The dataset used in their experiments included 124 training and 172 test images .
Their SPNet was trained for 30 epochs with a batch size of 50 images.

\begin{table*}[t]
\centering
\caption{Summary of influential papers appeared in 2016 and 2017 (based on Google Scholar citation index in January 2019) that exploit deep learning for the Segmentation tasks in Medical Image Analysis.} 
\label{tab:seg}
\begin{tabular}{|c|c|c|c|c|c|}
\hline
\textbf{Reference}    &\textbf{Anatomic Site}
& \textbf{Image Modality}     & \textbf{Network type}  & \textbf{Data}&
\textbf{Citations}
\\
\hline \hline
Milletari\textit{ et al.}~\cite{milletari2016v} (2016)          & Brain
&   MRI                 & FCN              &    Private data   &  550              \\ \hline
Ciceck \textit{ et al.}~\cite{cciccek20163d} (2016) &         Kidney         &  CT                & 3D U-Net                &         Private data    &  460                   \\ \hline
Pereira\textit{ et al.}~\cite{pereira2016brain} (2016)          & Brain
&   MRI                 &  CNN              &    BRATS2013, 2015    &  411   
\\ \hline
Moeskops\textit{ et al.}~\cite{moeskops2016automatic} (2016)          & Brain
&   MRI                 &  CNN              &   5 datasets   &  242   
\\ \hline
Liskowski  \textit{ et al.}~\cite{liskowski2016segmenting} (2016) &         Eye         &  Opthalmology                & DNN                &   400 images, DRIVE, STARE, CHASE          & 177                     \\ \hline
Ibragimov\textit{ et al.}~\cite{cheng2016computer} (2016)          & Breast
&   CT                & AE              &     1400 images   &  174             \\ \hline
Havaei \textit{ et al.}~\cite{havaei2017brain} (2017)&  Brain        
&      MRI
&       CNN             & 2013 BRATS  & 596         \\ \hline
Kamnistas\textit{ et al.}~\cite{kamnitsas2017efficient}(2017)          & Brain
&   MRI                 & 11 layers 3D-CNN              &     BRATS 2015 and ISLES 2015    &  517              \\ \hline

Fang\textit{ et al.}~\cite{fang2017automatic} (2017)          & Eye
&   OCT                & CNN             &     60 volumes   &  86           \\ \hline
Ibragimov\textit{ et al.}~\cite{ibragimov2017segmentation}(2017)          & Liver
&   CT                & CNN              &     50 images   &  56             \\ \hline

\end{tabular}
\end{table*}

Sarker \textit{et al.}~\cite{sarker2018slsdeep} analyzed  skin lesion segmentation with deep learning. They used autoencoder and decoder networks for feature extraction. The loss function is minimized in their work by combining negative Log Likelihood and end-point-error for the segmentation of melanoma regions with sharp edges. They evaluated their method SLSDeep on  ISBI datasets~\cite{codella2018skin}, \cite{gutman2016skin} for skin lesion detection, achieving encouraging segmentation results.
In another related study of skin cancer,  Mirikharaji \textit{et it.}~\cite{mirikharaji2018star} also developed a deep FCN for skin lesion segmentation. They presented good  results on ISBI 2017 dataset of dermoscopy images. They used two fully convolutional networks based on   U-Net~\cite{ronneberger2015u} and ResNet-DUC~\cite{he2016deep} in their technique. 
 Yuan \textit{et al.}~\cite{yuan2017automatic} proposed a deep fully convolutional-deconvolutional neural network (CDNN) for the automatic skin lesion segmentation, and acquired Jaccard index of 0.784 on the  validation set of ISBI. Ambellan \textit{ et al.} \cite{ambellan2018automated} proposed a CNN based on 3D Statistical Shape Models (SSMs) for the segmentation of knee and cartilage in MRI images. In Table~\ref{tab:seg}, we also summarize few popular methods in medical image segmentation that appeared prior to the year 2018.

\subsection{Registration}
\label{sec:reg}

Image registration is a common  task in medical image analysis that allows spatial alignment of images to a common anatomical space~\cite{klein2009evaluation}. 
It aims at aligning a source image with a target image through  transformations. 
Image registration is one of the main stream tasks in medical image analysis that has received ample attention  even before the deep learning era~\cite{bajcsy1989multiresolution,thirion1998image,beg2005computing,ashburner2007fast,avants2008symmetric,glocker2008dense,dalca2016patch}. 
Advent of deep learning has also caused neural networks to penetrate in medical image registration~\cite{krebs2017robust,rohe2017svf,sokooti2017nonrigid,yang2017quicksilver}. 

\subsubsection{Brain}
\label{sec:S_brain}
Van \textit{et al.}~\cite{van2018stacked} proposed a stacked bidirectional convolutional
LSTM (C-LSTM) network for the reconstruction of 3D images from the 4D spatio-temporal data. Previously,   \cite{nie2016estimating,bahrami2016convolutional} used CNN techniques for the reconstruction of 3D CT and MRI images using four 3D convolutional layers.
Lonning \textit{ et al.} ~\cite{lonning2018recurrent} presented a deep learning method using Recurrent Inference Machines (RIM) for the reconstruction of MRI.
Deep learning based deformable image registration has also  been recently performed by Sheikh  \textit{et al.}~\cite{sheikhjafari2018unsupervised}. They used deep FCN to generate spatial transformations through under feed forward networks. In their experiments,  they used  cardiac MRI images from ACDC 2017 dataset  and showed promising results in comparison to a moving mesh registration  technique. 
Hou \textit{ et al.}~\cite{hou2018deep} also proposed a learning based image registration technique using CNNs. They used 2D image slice transformation to construct 3D images using a canonical co-ordinate system. First, they simulated their approach on fetal MRI images and then used real fetal brain MRI images for the experiments. Their work is also claimed to be promising for computational efficiency. In another study of image based registration~\cite{hou2018image}, the same group of authors evaluated their technique on CT and MRI datasets for different loss functions using SE(3) as a  benchmark. They trained CNN directly on SE(3) and proposed a Riemannian manifold based  formulation for pose estimation problem.  The registration accuracy with their approach increased from 2D to 3D image based registration as compared to previous methods. The authors further showed in \cite{hou20183d} that CNN can reliably reconstruct 3D images using 2D image slices.  
Recently, Balakrishnan \textit{et al.}\cite{balakrishnan2018unsupervised} worked on 3D pairwise MR brain image registration. They proposed an unsupervised learning technique named VoxelMorph CNN. They used a pair of two 3D images as input, with dimensions  $160 \times 192 \times 224$; and learned shared parameters in their network for convolution layers. They demonstrated their method on 8 publicly available datasets of brain MRI images. On ABIDE dataset their model achieved $1.5$\% improvement in the dice score. It is claimed that their method is also  computationally more efficient than the exiting techniques for this problem.

\subsubsection{Eye}
\label{sec:S_eye}
Costa \textit{et al.}~\cite{costa2018end} used adversarial autoencoders for the synthesis of retinal colored images. They trained a generative model to generate synthetic images  and another model to classify its output into a real or synthetic. The  model results in an end-to-end retinal image synthesis system and generates as many  images as required by its users.
It is demonstrated that the image space learned by the model has an arguably well defined semantic structure. 
The synthesized images were shown to be visually and quantitatively different from the images used for training their model. The shown images reflect good visual quality. Mahapatra \textit{et al.}~\cite{mahapatra2018deformable} proposed an end-to-end deep learning method using generative adversarial networks for  multimodal image registration. They used retinal and cardiac images for registration. Tang \textit{et al.}~\cite{tang2018retinal} demonstrated a robust image registration approach based on mixture feature and structure preservation (MFSP) non rigid point matching method. In their method they first  extracted feature points by speed up robust feature (SURF) detector and partial intensity invariant feature descriptor (PIIFD) from model and  target retinal image. Then they used MFSP for feature map detection.    

Pan \textit{et al.}~\cite{pan2018detection} developed a deep learning technique to remove eye position difference for longitudinal 3D retinal OCT images. In their method, they first perform pre-processing for projection image then, to detect vessel shadows, they apply enhancement filters. The SURF algorithm~\cite{bay2006surf} is then used to extract the  feature points, whereas RANSAC~\cite{fischler1981random}  is applied for cleaning the outliers. 
Mahapatra \textit{et al.}~\cite{mahapatra2018elastic} also proposed an end-to-end deep learning technique for image registration. They used GANs which registered images in a single pass with deformation field. They used ADAM optimizer~\cite{kingma2014adam} for minimizing the network cost and trained the model using the Mean Square Error (MSE) loss. 

\subsubsection{Chest}
\label{sec:S_chest}
Relating to the anatomical region of chest, Eppenhof \textit{et al.}~\cite{eppenhof2018deformable} proposed a 3D FCN based technique for the registration of CT lung inspiration-expiration image pairs. They validated the performance of their method using two datasets, namely  DIRLAB~\cite{castillo2009four} and CREATIS~\cite{vandemeulebroucke2012automated}. In general, there is a growing perception in the Medical Imaging community that Deep learning is a promising tool for 2D and 3D image registration for the chest regions. 
De\textit{ et al.}~\cite{de2019deep} also  trained a CNN model for the affine and deformable image registration. Their technique allows registration of the pairs of unseen images in a single pass. They applied their technique to cardiac cine MRI and chest CT images for registration.
Zheng \textit{et al.}~\cite{zheng2018pairwise} trained a CNN model for 2D/3D image registration problem under a Pairwise Domain Adaptation (PDA) technique that uses synthetic data. It is claimed that their method can learn effective representations for image registration  with only  a limited number of training images. They demonstrated   generalization and flexibility of their method for clinical applications. Their PDA method can be specially suitable where small training data is available.

\subsubsection{Abdomen}
\label{sec:S_abdomen} 
Lv \textit{ et al.}~\cite{lv2018respiratory} proposed a CNN based technique for the 3D MRI abdomen image registration.
They trained their model for the spatial transformation analysis of different images. To demonstrate the effectiveness of their technique, they compared their method with three other approaches and claimed a reduction in the reconstruction time from 1h to 1 minute. In another related study, Lv\textit{ et al.}~\cite{lv2018performance} proposed a deep learning framework based on the popular U-net architecture. 
To evaluate the performance of their technique they used 8 ROI's from cortex and medulla of segmented kidney. It is demonstrated by the authors that during free breathing measurements, their normalized root-mean-square error (NRMSE) values for cortex  and medulla were significantly lower after registration. 

\subsubsection{Miscellaneous }
\label{sec:S_misc}

Yan \textit{ et al.}~\cite{yan2018adversarial} presented an Adversarial Image Registration (AIR) method for multi-modal image MR-TRUS registration~\cite{hu2018label}. They trained two deep networks concurrently, one for the generator component of the adversarial framework, and the other for the discriminator component. In their work, the authors learned not only an image registration network but also a so-called metric network which computes the quality of image registration.
The data used in their experimentation consists of 763 sets of $3$D TRUS volume and $2$D MR volume with $512$x$512$x$26$ voxels. The developed AIR network is also  evaluated on clinical datasets acquired through image-fusion guided prostate biopsy procedures.
For the visualization of 3D medical image data Zhao \textit {et al.}~\cite{zhao2018respond} recently proposed a deep learning based technique,  named Respond-weighted Class Activation Mapping (Respond-CAM). As compared to Grade-CAM~\cite{selvaraju2017grad} they claim better performance. 
Elss \textit{et al.}~\cite{elss2018motion} also employed Convolutional networks for  single phase image motion in cardiac CT 2D/3D images. They trained regression network to successfully learn 2D motion estimation vectors. We also summarize few worth noting contributions from the years 2016 and 2017 in Table~\ref{tab:reg}.

\begin{table*}[t]
\centering
\caption{Summary of influential papers appeared in 2016 and 2017 (based on Google Scholar citation index in January 2019) that exploit deep learning for the Registration task in Medical Image Analysis.} 
\label{tab:reg}
\tabcolsep=0.08cm
\begin{tabular}{|c|c|c|c|c|c|}
\hline
\textbf{Reference}    &\textbf{Anatomic Site}
& \textbf{Imaging Modality}     & \textbf{Network type}  & \textbf{Data}&
\textbf{Citations}
\\
\hline \hline
Miao\textit{ et al.}~\cite{miao2016cnn} (2016) &         Chest       &  X-ray                & CNN regression                &      Synthetic      &  101                  \\ \hline
Wu\textit{ et al.}~\cite{wu2016scalable} (2016)          & Brain
&               MRI    & CNN              &    LONI, ADNI databases &  58            \\ \hline
Simonvosky\textit{ et al.}~\cite{simonovsky2016deep} (2016)          & -
&  Multiple   & CNN              & 
{IXI} &  57            \\ \hline
Yang\textit{ et al.}~\cite{yang2016fast} (2016)          & Brain
&               MRI    & Encoder-decoder              &   OASIS dataet &  40            \\ \hline
Barahmi\textit{ et al.}~\cite{bahrami2016convolutional} (2016)          & Brain
&               MRI    & CNN             &   15 subjects &  36            \\ \hline
Zhang\textit{ et al.}~\cite{zhang2016deep} (2016)          & Head, Abdomen, Chest
&               CT    & CNN
              &   Private  &  30            \\ \hline
Nie\textit{ et al.}~\cite{nie2017medical} (2017)          & Multi task
&               MRI, CT    & FCN
             &Private  &  110           \\ \hline
Kang\textit{ et al.}~\cite{kang2017deep} (2017)          & Abdomen
&        CT          & CNN & CT low-dose Grand Challenge  &  98            \\ \hline
Yang\textit{ et al.}~\cite{yang2017quicksilver} (2017)          & Brain
&               MRI    & Encoder-decoder              &   373 OASIS and 375 IBIS images &  64            \\ \hline
De\textit{ et al.}~\cite{de2017end} (2017)          & Brain
&               MRI    & CNN              &   Sunnybrook Cardiac Data~\cite{radau2009evaluation} &  50           \\ \hline

\end{tabular}
\end{table*}

\subsection{Classification}
\label{sec:class}
Classification of images is a long standing problem in Medical Image Analysis and other related fields, e.g.~Computer Vision. In the context of medical imaging, classification becomes a fundamental task for Computer Aided Diagnosis (CAD). Hence, it is no surprise that many researchers have recently tried to exploit the advances of deep learning for this task in medical imaging.

\subsubsection{Brain}
\label{sec:C_brain}
Relating to the anatomical region of Brain, Li {\it et al.}~\cite{li2018brain}  used deep learning to detect Autism Spectrum disorder (ASD) in functional Magnetic Resonance Imaging (fMRI).
They developed a 2-stage neural network method. For the first stage, they trained a CNN (2CC3D) with 6 convolutional layers, 4 max-pooling layers and 2 fully connected layers. Their network uses a   sigmoid output layer. For the second stage, in order to detect biomarkers for ASD, they took advantage of the anatomical structure of brain fMRI. They  developed a frequency normalized sampling method for that purpose. Their method is evaluated using  multiple databases, showing robust results for  neurological function of biomarkers. 

In their work, Hosseini-Asl \textit{et al.}~\cite{aslalzheimer} employed an auto-encoder  architecture for diagnosing Alzheimer's Disease (AD) patients. They reported up to $99\%$ accuracy on {ADNI dataset}. They exploited Transfer Learning to handle the data scarcity issue, and used a model that is pre-trained with the help of {CAD Dementia dataset}. Their network architecture is based on 3D convolutional kernels that models generic brain features from sMRI data. 
The overall classification process in their technique first spatially normalizes the brain sMRI data, then it learns the 3D CNN model using the normalized data. The model is eventually fine-tuned on the target domain, where the fine-tuning is performed in a supervised manner. 

Recently, Yan \textit{et al.}~\cite{yan2018deep}  proposed a deep chronectome learning framework for the classification of MCI in brain using Full Bidirectional Long Short-Term Memory (Full-BiLSTM) networks. Their method can be divided into two parts, firstly a Full-LSTM is used to gather time varying information in brain for which MCI can be diagnosed. Secondly, to mine the contextual information hidden in dFC, they applied BiLSTM to access long range context in both directions.
They reported the performance of their model on public dataset  ADNI-2,
achieving  $73.6\%$ accuracy. 
Hensfeld \textit{et al.}~\cite{heinsfeld2018identification} also proposed a deep learning algorithm for the Autism Spectrum disorder (ASD) classification in rs-fMRI images on multi-site database ABIDE. They used denoising autoencoders for unsupervised pretraining. The classification accuracy achieved by their algorithm on the said dataset is~$70\%$.

In the context of classification related to the anatomical region of brain, Soussia {\it et al.}~\cite{soussia2018review} provided a review of 28 papers from 2010 to 2016 published in MICCAI. They reviewed neuroimaging-based technical methods developed for the Alzheimer Disease (AD) and Mild-Cognitive Impairment (MCI) classification tasks. 
The majority of papers used MRI for dementia classification and few worked to predict MCI conversion to AD at later observations. We refer to~\cite{soussia2018review} for the detailed discussions on the contributions reviewed by this article.  
Gutierrez \textit{et al.}~\cite{gutierrez2018deep} proposed a deep neural network, termed  Multi-structure point network (MSPNet), for the shape analysis on multiple structures. This network is  inspired by PointNet~\cite{qi2017pointnet} that can directly process point clouds. MSPNet achieves good  classification accuracy for AD and MCI for the  ADNI database.

\subsubsection{Breast}
\label{sec:C_breast}
Awan {\it et al.}~\cite{awan2018context} proposed to use more context information for breast image classification. They used features of a CNN that is  pre-trained on ICIAR 2018 dataset for histological images~\cite{araujo2017classification}, and classified breast cancer as benign, carcinoma insitu (CIS) or breast invasive carcinoma (BIC). 
Their technique performs patch based and context aware image classification. They used ResNet50 architecture and overlapping patches of size $512 \times 512$. The extracted features are classified using a Support Vector Machine in their approach. Due to the unavailability of large-scale data, they used random rotation and flipping data augmentation techniques during the training process.
It is claimed that their trained model can also be applied to other tasks where contextual information and high resolution are required for optimal prediction.

When only weak annotations are available for images, such as in heterogeneous images, it is often useful to turn to multiple instance  learning (MIL). Courture \textit{et al.}~\cite{couture2018multiple} described a CNN using quantile function for the classification of 5 types of breast tumor histology. They fine-tuned AlexNet~\cite{krizhevsky2012imagenet}.  
The data used in their study consists of 1,713 images from the Carolina Breast Cancer Study, Phase 3~\cite{troester2018racial}. They improved the classification accuracy on this dataset  from 68.6 to 85.6 for estrogen receptor (ER) task in breast images. Recently, MIL has also been used for breast cancer classification in \cite{sudharshan2019multiple} and \cite{roy2019patch} that perform patch based classification of histopathology images. 
Antropova \textit{et al.}~ \cite{antropova2018use} used 690 cases with 5 fold cross-validation of MRI maximum intensity projection for breast lession classification. They used a pre-trained VGGNet~\cite{simonyan2014very} for feature extraction, followed by an SVM classifier. 
Ribli \textit{et al.}~\cite{ribli2018detecting} applied a CNN based on VGG16 for lession classification in mammograms. They trained their model using DDSM dataset and tested it on INbreast~\cite{moreira2012inbreast}. They achieved the second best score for \href{http://sagebionetworks.org/research-projects/digital-mammography-dream-challenge/}{Mammography DREAM Challenge}, with AUC of $0.95$. 
Zheng \textit{et al.}~\cite{zheng2018breast} proposed a CAD technique for breast cancer classification using CNN based on pre-trained VGG-19 model. They evaluated their technique's performance on digital mammograms of pairs of 69 cancerous and 27 healthy subjects. They achieved the values of  $0.928$ for sensitivity and $0.991$ for specificity of classification.

\subsubsection{Eye}
\label{sec:C_eye}
Pertaining to the region of eye, Gergeya \textit{et al.}~\cite{gargeya2017automated} took a data driven approach using deep learning to classify 
Diabetic retinopathy (DR) in color fundus images. The authors used public databases MESSIDOR 2 and E-ophtha to train and test their models and achieved $0.94$ and $0.95$ AUC score respectively on the test partitions of these datasets. 
A convolutional network is also employed by Pratt \textit{et al.}~\cite{pratt2016convolutional} for diagnosing and classifying the severity of DR in color fundus images. Their model is trained using the Kaggle datasets, and it achieved $75\%$ DR severicity accuracy.
Similarly, Ayhan \textit {et al.}~\cite{ayhan2018test} also exploited the deep CNN architecture of ResNet50~\cite{he2016identity} for the fundus image classification. 
Mateen \textit{et al.}~\cite{mateen2019fundus} proposed a DR classification system based on VGG-19. They also  performed evaluation using  Kaggle dataset of 35,126 fundus images. It is claimed that their model outperforms the more conventional techniques, e.g. SIFT as well as earlier deep networks, e.g. AlexNet in terms of accuracy for the same task. 

\subsubsection{Chest}
\label{sec:C_chest}
Dey {\it et al.}~\cite{dey2018diagnostic} studied 3D CNNs for the diagnostic classification of lung cancer between benign and malignant in CT images. Four networks were analyzed for their classification task, namely a basic 3D CNN; a  multi-output CNN; a 3D DenseNet, and an augmented 3D DenseNet with multi-outputs. They employed the public dataset LIDC-IDRI with $1,010$ CT images and a private dataset of $47$ CT images with both malignant and benign in this study. 
The best results are achieved by the 3D multi-output DenseNet (MoDenseNet) on both datasets, having accuracy $90.40\%$ as compared to previously reported accuracy of $89.90\%$~\cite{nibali2017pulmonary}. 
Gao \textit {et al.}~\cite{gao2018holistic}  proposed a deep CNN for the classification of Interstitial Lung Disease (IDL) patterns on CT images. Previously,  batch based algorithms were being used for this purpose~\cite{song2015large,song2013feature}.
In contrast, Gao \textit {et al.} performed holistic classification using the complete image as network input. Their experiments used a  public data~\cite{depeursinge2012building} on which the classification accuracy improved to $87.9\%$ from the previous results of $86.1\%$~\cite{song2013feature}.
For the holistic image classification, the overall accuracy of $68.8\%$ was achieved. In another work, Hoo \textit{et al.}~\cite{hoo2016deep} et al. also analyzed three CNN architectures, namely CifarNet, AlexNet, and GoogLeNet, for interstitial lung disease classification.

Biffi \textit{et al.}~\cite{biffi2018learning} proposed a 3D convolutional generative model for the classification of cardiac diseases. They achieved impressive performance for the classification of healthy and hypertrophic cardiomyopathy MR images. For ACDC MICCAI 2017 dataset they were able to achieve 90\% accuracy for classification of healthy subjects. Chen \textit{et al.} \cite{brestel2018radbot} proposed a CNN based technique RadBot-CXR to categorize focal lung opacities, diffuse lung opacity, cardiomegaly, and abnormal hilar prominence in chest X-ray images. They claim that their algorithm showed radiologists level performance for this task.  
Wang \textit{ et al.}~ \cite{wang2018weakly} used deep learning in analyzing histopathology images for the whole slide lung cancer classification. 
Coudray\textit{ et al.}~\cite{coudray2018classification} used Inception3 CNN model to analyze whole slide images to classify lung cancer between  adenocarcinoma (LUAD), squamous cell carcinoma (LUSC) or normal tissues. 
Moreover, they also trained their model for the prediction of ten most common mutated genes in LUAD and achieved good accuracies. 
Masood \textit{et al.}~\cite{masood2018computer} proposed a deep learning approach DFCNet based on FCN, which is used to classify the four stages of detected pulmonary lung cancer nodule. 

\subsubsection{Abdomen}
\label{sec:C_abdomen}
Relating to abdomen, Tomczak \textit{ et al.}~\cite{tomczak2018histopathological} employed deep Multiple Instance Learning (MIL)  framework~\cite{ilse2018attention} for the classification of esophageal cancer in histopathology images.   
In another contribution, Frid \textit{et al.}~\cite{frid2018synthetic} used GANs for the synthetic medical image data generation. 
They made classification of CT liver images as their test bed and performed classification of 182 lesions. The authors demostrated that by using augmented data with the GAN framework, upto $7\%$ improvement is possible in classification accuracy.  
For automatic classification of ultrasound abdominal images, Xu \textit{et al.}~\cite{xu2018less} proposed a multi-task learning framework based on CNN. For the experiments they used 187,219 ultrasound images and claimed better classification accuracy than human clinical experts.

\subsubsection{Miscellaneous }
\label{sec:C_misc}

Esteva \textit{et al.}~\cite{esteva2017dermatologist} presented a  CNN model to classify skin cancer lesions. Their model is trained in an end-to-end manner   directly from the images, taking pixels and disease labels as inputs. They used datasets of 129,450 clinical images consisting of 2,032 different diseases to train CNN. They classified two most common cancer diseases; keratinocyte carcinomous versus benign seborrheic keratosis, and the deadliest skin cancer; malignant melanomas versus benign nevi.
For skin cancer sun exposure classification, Combalia \textit{ et al.} ~\cite{combalia2018monte} also applied a Monte Carlo method to only highlight the most elistic regions. Antony \textit{et al.}~\cite{antony2016quantifying} employed a deep CNN model for automatic quantification of severity of knee osteoarthritis (OA). They used Kellgren and  Lawrence (KL) grades to assess the severity of knee. 
In their work, using deep CNN pre-trained on ImageNet and fine-tuned on knee OA images resulted in good  classification performance. Paserin \textit{et al.}~\cite{paserin2018real} recently worked on diagnosis and classification of developmental dysplasia of hip (DDH). They proposed a CNN-RNN technique for 3D ultrasound volumes for DDH. Their model consists of convolutional layers for feature learning  followed by recurrent layers for spatial relationship of their responses. Inspired by VGG network~\cite{simonyan2014very}, they used CNN with 5 convolutional layers for feature extraction with ReLU activations, and $2\times2$ max-pooling with a stride of 2. Finally, they used LSTM network that has 256 units. They achieved $82$\% accuracy with AUC $0.83$ for 20 test volumes.
Few notable contributions from the years 2016 and 2017 are also summarized in Table~\ref{tab:class}.

\begin{table*}[t]
\centering
\caption{Summary of notable contributions appearing in 2016 and 2017 that exploit deep learning for the Classification task in Medical Image Analysis. The citation index is based on Google Scholar (January 2019).} 
\label{tab:class}
\tabcolsep=0.08cm
\begin{tabular}{|c|c|c|c|c|c|}
\hline
\textbf{Reference}    &\textbf{Anatomic Site}
& \textbf{Image Modality}     & \textbf{Network type}  & \textbf{Data}&
\textbf{Citations}
\\
\hline \hline

Anthimopoulos\textit{ et al.}~\cite{anthimopoulos2016lung} (2016) &         Lung        &  CT                & CNN                 &   Uni.~ Hospital of Geneva~\cite{depeursinge2012building}\& Inselspital      
&  253                   \\ \hline
Kallenberg\textit{ et al.}~\cite{kallenberg2016unsupervised} (2016) &         Breast      &  Mammography                & CNN               &         3 different databases    &  111                   \\ \hline
Huynh\textit{ et al.}~\cite{huynh2016digital} (2016) &         Breast      &  Mammography                & CNN               &         Private data (219 lesions)    &  76                   \\ \hline
Yan\textit{ et al.}~\cite{yan2016multi} (2016) &         12 regions     &  CT                & CNN               &         Synthetic \& Private    &  73                   \\ \hline
Zhang\textit{ et al.}~\cite{zhang2016deep} (2016) &         Breast      & Elastography                 & MLP               &         Private data (227 images)   &  50                   \\ \hline
Esteva\textit{ et al.}~\cite{esteva2017dermatologist} (2017)          & Skin
&               Histopatholohy    & CNN              &    Private large-scale &  1386             \\ \hline
Sun\textit{ et al.}~\cite{sun2017enhancing} (2017)          & Breast
&       Mammography   & CNN              &    Private data &  40             \\ \hline
Christodoulidis\textit{ et al.}~\cite{christodoulidis2016multi} (2017)          & Lung
&               CT    & CNN              & Texture data as Transfer learning source &  36             \\ \hline

Lekadir\textit{ et al.}~\cite{lekadir2017convolutional} (2017)          & Heart
&               US    & CNN             &    Private data  &  26             \\ \hline
Nibali\textit{ et al.}~\cite{nibali2017pulmonary} (2017)          & Lung
&               CT   & CNN              &    LIDC/IDRI dataset  &  22             \\ \hline

\end{tabular}
\end{table*}

\label{sec:det}

\section{Datasets}
\label{sec:Dataset}
Good quality data has always remained the primary requirement for learning reliable computational models.  This is also true for deep models that also have  the additional requirement of consuming large amount of training data. Recently, many public datasets for medical imaging tasks have started to emerge. There is also a  growing trend in the research community to compile lists of these datasets. For instance we can find few useful compilation of public dataset lists at 
\href{https://github.com/sfikas/medical-imaging-datasets}{Github repositories} and other \href{http://homepages.inf.ed.ac.uk/rbf/CVonline/Imagedbase.htm}{webpages}. Few medical image analysis \href{http://www.osirix-viewer.com/resources/dicom-image-library/}{products} are also helping in providing public datasets. Whereas detailed discussion on the currently available public datasets for medical imaging tasks is outside the scope of this article, we provide typical examples of the commonly used datasets in medical imaging by deep learning approaches in Table~\ref{tab:data}.
The Table is not intended to provide an exhaustive list. We recommend the readers  internet search for that purpose. A brief search can result in a long compilation of medical imaging datasets.
However, we summarize few examples of contemporary datasets in Table~\ref{tab:data} to make an important point regarding deep learning research in the context of Medical Image Analysis.
With the exception of few datasets, the public datasets currently available for medical imaging tasks are small in terms of the number of samples and patients.  As compared to the datasets for general Computer Vision problems, where datasets typically range from few hundred thousand to millions of annotated images, the dataset sizes for Medical imaging tasks are too small.
On the other hand, we can see the emerging trend  in Medical Imaging community of adopting the practices of broader Pattern Recognition community, and aiming at learning deep models in end-to-end fashion.  However, the broader community has generally adopted such practices based on the availability of large-scale annotated datasets, which is an important requirement for inducing reliable deep models.
Hence, it remains to be seen that how effectively  end-to-end trained models can really perform the medical image analysis tasks without over-fitting to the training datasets.

\begin{table*}[t]
\centering
\caption{Examples of popular databases used by Medical Image Analysis techniques that exploit Deep Learning.} 
\label{tab:data}
\begin{tabular}{|c|c|c|c|c|c|}
\hline
\textbf{Sr.}    &\textbf{Database}
& \textbf{Anatomic site}     & \textbf{Image modality}    & \textbf{Main task}    & \textbf{Patients/Images}   \\
\hline \hline
1               & \href{ http://medgift.hevs.ch/wordpress/databases/ild-database/}{ILD}~\cite{depeursinge2012building} 
& Lung                   & CT                 &  Classification                  &  120 patients                    \\ \hline
2               & \href{https://wiki.cancerimagingarchive.net/display/Public/LIDC-IDRI}{LIDC-IDRI}\cite{armato2011lung}                  & Lung                  & CT                 & Detection                   &  1,018  patients                   \\ \hline
3               & \href{https://www.ncbi.nlm.nih.gov/pubmed/23110865}{ADNI}~\cite{petersen2010alzheimer}                  & Brain                  & MRI                 & Classification                  & >800 patients                       \\ \hline
4               & \href{https://stanfordmlgroup.github.io/competitions/mura/}{MURA}~\cite{rajpurkar2017mura}                   & Musculoskeletal                   & X-ray                 & Detection                    & 40,561  images                    \\ \hline
5               & \href{http://www.cancerimagingarchive.net/}{TCIA }                  &  Multiple                 & Multiple                 &  Multiple                  & >35,000 patients                      \\ \hline
6               & \href{https://www.med.upenn.edu/sbia/brats2018/data.html}{BRATS}~\cite{menze2015multimodal}                  &  Brain             &      MRI            &      Segmentation              &  -                     \\ \hline
7               & \href{http://www.eng.usf.edu/cvprg/Mammography/Database.html}{DDSM}~\cite{heath2000digital}                   &         Breast          &           Mamography       &           Segmentation         &     2,620 patients\\ \hline 
8              & \href{http://latim.univ-brest.fr/indexfce0.html}{MESSIDOR-2}~\cite{quellec2008optimal},\cite{decenciere2014feedback}
                  & Eye              & OCT                  & Classification         & 1,748 images                      \\ \hline
9               & \href{https://www.nih.gov/news-events/news-releases/nih-clinical-center-provides-one-largest-publicly-available-chest-x-ray-datasets-scientific-community}{ChestX-ray14}~\cite{wang2017chestx}                    & Chest                    & X-ray                 & Multiple                   & >100,000 images                       \\ \hline
10  &  \href{https://www.creatis.insa-lyon.fr/Challenge/acdc/databases.html}{ACDC 2017}                  & Brain                  & MRI                &  Classification                  & 150 patients                    \\ \hline
11  & \href{https://warwick.ac.uk/fac/sci/dcs/research/tia/glascontest/}{2015 MICCAI Gland Challenge}                & Glands                 & Histopathology               &  Segmentation                  & 165 images                     \\ \hline       
12  &  \href{https://data-archive.nimh.nih.gov/oai/}{OAI}                &  Knee                 & X-ray, MRI                &         Multiple           & 4,796 patients                     \\ \hline
13  & \href{http://homes.esat.kuleuven.be/~mblaschk/projects/retina/}{DRIVE}~\cite{staal2004ridge},\cite{niemeijer2004comparative}                &  Eye                 & SLO               &  Segmentation                  & 400  patients                     \\ \hline
14  & \href{http://homes.esat.kuleuven.be/~mblaschk/projects/retina/}{STARE}~\cite{hoover2000locating}                &  Eye                 & SLO               &  Segmentation                  & 400 images                      \\ \hline
15  & \href{https://blogs.kingston.ac.uk/retinal/chasedb1/}{CHASEDB1}~\cite{fraz2012ensemble}                &  Eye                 & SLO               &  Segmentation                  & 28 images \\ 
\hline
16 & \href{https://www.oasis-brains.org/}{OASIS-3}~\cite{fotenos2005normative,morris1993clinical, buckner2004unified,rubin1998prospective,zhang2001segmentation,marcus2010open}, \cite{marcus2007open}              &  Brain                 & MRI, PET              &  Segmentation                  &        1,098 patients              \\
\hline
17&     \href{http://www.mammoimage.org/databases/}{MIAS} ~\cite{suckling2015mammographic}          &  Breast                 & Mammography             &  Classification                  &        322 patients          \\
\hline
18 &  \href{http://www.isles-challenge.org/}{ISLES 2018}           &    Brain                & MRI             & Segmentation                  &        103 patients          \\
\hline
19&  \href{http://segchd.csail.mit.edu/}{HVSMR 2018}~\cite{pace2015interactive}           &         Heart          & CMR             & Segmentation                  &        4 patients      \\
\hline
20&  \href{https://camelyon17.grand-challenge.org/}{CAMELYON17}~\cite{bandi2018detection}          &         Breast         & WSI             & Segmentation                  &        899 images      \\
\hline
21&  \href{https://challenge2018.isic-archive.com/}{ ISIC 2018}           &         Skin         & JPEG             & Detection                  &        2,600 images       \\
\hline
22&  \href{https://openneuro.org/}{OpenNeuro}           &         Brain       & Multiple             &         Classification         &        4,718 patients      \\
\hline
23&  \href{http://fcon_1000.projects.nitrc.org/indi/abide/}{ ABIDE}           &         Brain       & MRI            &            Disease Diagnosis     &       1,114 patients       \\
\hline

24&  \href{http://medicalresearch.inescporto.pt/breastresearch/index.php/Get_INbreast_Database}{ INbreast}~\cite{moreira2012inbreast}           &         Breast       & Mammography           &    Detection/Classification   &     410 images   \\
\hline
\end{tabular}
\end{table*}

\section{Challenges in Going Deep}
\label{sec:chal}
In this Section, we discuss  the major challenges faced in fully exploiting the powers of Deep Learning in Medical Image Analysis. Instead of describing the issues encountered in specific tasks, we focus more on the fundamental challenges and explain the root causes of these problems for the Medical Imaging community that can also help in understanding the task-specific challenges. Dealing with these challenges is the topic of discussion in  Section~\ref{sec:FD}.

\paragraph{Lack of appropriately annotated data}
It can be argued that the single aspect  of Deep Learning that sets it apart from the rest of  Machine Learning techniques is its ability to model extremely complex mathematical functions. Generally, we introduce more layers to learn more complex models - i.e.~go deep.
However, a deeper network must also learn more model parameters. A model with a large number of parameters can only generalize well if we correspondingly use a very large amount of data to infer the parameter values. 
This phenomenon is fundamental to any Machine Learning technique. A complex model inferred using a limited amount of data normally over-fits to the used data and performs poorly on any other data. Such modeling is highly undesirable because it gives a false impression of learning the actual data distribution whereas the model is only learning the peculiarities of the used training data. 

Learning deep models is inherently unsuitable for the domains where only limited amount of training data is available. Unfortunately, Medical Imaging is one such domain. For most of the problems in Medical Image Analysis, there is only a limited amount of data that is annotated in a manner that is suitable to learn powerful deep models. We encounter the problem of `lack of appropriately annotated data' so frequently in the current Deep Learning related Medical Imaging literature that it is not difficult to single out this problem as `the fundamental challenge' that Medical Imaging community is currently facing in fully exploiting the advances in Deep Learning.  

The Computer Vision community has been able to take full advantage of Deep Learning because data annotation is relatively straightforward in that domain. Simple crowd sourcing can yield millions of accurately annotated images. This is not possible for Medical Images that require high level of specific expertise for annotation. Moreover, the stakes are also very high due to the nature of medical application, requiring extra care in annotation. Although we can also find large number of images in medical domain via systems like PACS and OIS, however using them to train deep models is still not easy because they lack appropriate level of annotations that is generally required for training useful deep models.

With only a few exceptions, e.g.~\cite{wang2017chestx} the public datasets available in the Medical Imaging domain are not  large-scale - a requirement for training effective deep models. In addition to the issues of hiding patient's privacy, one major problem in forming  large-scale public datasets is that the concrete labels required for computational modeling can often not be easily inferred from medical reports. This is problematic if inter-observers are used to create  large-scale datasets. Moreover, the required annotations for deep models often do not perfectly align with the general medical routines. This becomes an additional problem even for the experts to provide noise-free annotations. 

Due to the primary importance of large-scale training datasets in Deep Learning there is an obvious need to develop such public datasets for Medical Imaging tasks. However, considering the practical challenges in  accomplishing this goal it is also imperative to simultaneously develop techniques of exploiting Deep Learning  with less amount of  data. We provide discussion on future directions along both of these dimension in Section~\ref{sec:FD}.


\paragraph{Imbalanced data}
One problem that occurs much more commonly in Medical Imaging tasks as compared to general Computer Vision tasks is the imbalance of samples in datasets.
For instance, a dataset to train a model for detecting breast cancer in mammograms may contain only a limited number of positive samples but a very large number of negative samples. Training deep networks with imbalanced data can induce models that are biased. Considering the low frequency of occurrences of positive samples for many Medical Imaging tasks, balancing out the original data can become as hard as developing large-scale dataset.  Hence, extra care must be taken in inducing deep models for Medical Imaging tasks.  

\paragraph{Lack of confidence interval}
Whereas the Deep Learning literature often refers to the output of a model as `prediction confidence'; the output signal of a neuron can only be interpreted as a \textit{single} probability value. The lack of provision of confidence interval around a predicted value is generally not desirable in Medical Imaging tasks. Litjens et al.~\cite{litjens2017survey} has noted that an increasing number of deep learning methods  in Medical Imaging are striving  to learn deep models in an end-to-end manner. Whereas end-to-end learning is the epitome of Deep Learning, it is not certain if this is the right way to exploit this technology in Medical Imaging. To an extent, this  conundrum is also hindering the widespread use of Deep Learning in Medical Imaging.

\section{Future directions}
\label{sec:FD}
With the recent increasing trend of exploiting Deep Learning in Medical Imaging tasks, we are likely to see a large influx of papers in this area in the near future. Here, we provide  guidelines and directions to help those works in dealing with the inherent challenges faced by Deep Learning in Medical Image Analysis. We draw our insights from the reviewed literature and the  literature in the sister fields of Computer Vision, Pattern Recognition and Machine Learning. Due to the earlier use of Deep Learning in those fields, the techniques of dealing with the related challenges have considerably matured in those areas. Hence, Medical Image Analysis can readily benefit from those findings in setting fruitful future directions.

Our discussion in this Section is primarily  aimed at providing guiding principles for the Medical Imaging community. Therefore, we limit it to the fundamental issues in Deep Learning. Based on the challenges mentioned in the preceding Section, and the insights from the parallel scientific fields, we present our discussion along three directions, addressing the following major questions. (1)~How can Medical Image Analysis still benefit from Deep Learning in the absence of large-scale annotated datasets? (2)~What can be done for developing large-scale Medical Imaging  datasets.  (3)~What should be the broader outlook of this research direction to catapult it in taking full advantage of the advances in Deep Learning?        


\subsection{Dealing with smaller data size}
\label{sec:SmallData}
\subsubsection{Disentangling Medical Task Transfer Learning}
\label{sec:task}
Considering the obvious lack of large-scale annotated datasets, Medical Imaging community has already started exploiting  `transfer learning'~\cite{azizi2017transfer}, \cite{kooi2017discriminating}, \cite{samala2017multi}.
In transfer learning, one can learn a complex model using data from a \textit{source} domain where large-scale annotated images are available (e.g.~natural images). Then, the model is further fine-tuned with  data of the \textit{target} domain where only a small number of annotated images are available (e.g.~medical images). It is clear from the literature that transfer learning is proving advantageous for Medical Image Analysis. Nevertheless, one promising recent development in transfer learning~\cite{zamir2018taskonomy} in Computer Vision literature remains completely unexplored for Medical Image Analysis.

Zamir et al.~\cite{zamir2018taskonomy} recently showed that performance of transfer learning can be improved by carefully selecting the source and target domains/tasks. By organizing different tasks that let the deep models  transfer well between themselves, they developed a so-called `taskonomy' to guide the use of transfer learning for natural images. This concept has received significant appreciation in the Computer Vision community, resulting in the `best paper' award for the authors at the prestigious IEEE Conference on Computer Vision and Pattern Recognition (CVPR), 2018. A similar concept is worth exploring for the data deprived Medical Imaging tasks. Disentangling medical tasks for transfer learning may prove very beneficial. Another related research direction that can help in dealing with smaller data size is to quantify the suitability of transfer learning between medical imaging and natural imaging tasks. A definitive understanding of the knowledge transfer abilities of the existing natural image models to the medical tasks can have a huge impact in Medical Image Analysis using Deep Learning.  


\subsubsection{Wrapping Deep Features for Medical Imaging Tasks}
\label{sec:wrap}
The existing literature shows an increasing trend of training deep models for Medical tasks in an `end-to-end' manner. For Deep Learning, end-to-end modeling is generally more promising for the domains where large-scale annotated data is available. Exploiting the existing deep models as feature extractors and then performing further learning on those features is a much more promising direction in the absence of large-scale training datasets. There is a considerable evidence in the Pattern Recognition literature  that activation signals of deeper layers in neural networks often form highly expressive image features. For natural images, Akhtar et al.~\cite{akhtar2017joint}  demonstrated that features extracted from deep models can be used to learn further effective higher level features using the techniques that require less training samples. They used Dictionary Learning framework~\cite{mairal2009online} to further wrap the deep features before using them with a classifier.

We note that Medical Image Analysis literature has already seen reasonably  successful attempts of using the existing natural image deep models as feature extractors e.g.~\cite{bar2015deep}, \cite{ciompi2015automatic}, \cite{lopes2017pre}. However, such attempts generally directly feed the features extracted from the pre-trained models to a classifier. The direction we point towards  entail  post-processing of deep features to better suit the requirements of the underlying Medical Image Analysis task.  

\subsubsection{Training Partially Frozen Deep Networks}
\label{sec:partial}
As a general principle in Machine Learning, more amount of training data is required to train more complex computational models. In Deep Learning, the network depth normally governs the model complexity, whereas deeper networks also have more parameters that require large-scale datasets for training. It is known  that the layers of CNNs - the most relevant neural networks for image analysis - systematically break down images into their features from lower level of abstraction to higher level of abstraction~\cite{yosinski2015understanding}. It is also known that the initial layers of CNNs learn very similar filters for a variety of natural images. These observations point towards the possibility of reducing the number of learn-able parameters in a network by freezing few of its layers to the parameter values that are likely to be similar for a variety of images. Those parameter values can be directly  borrowed from other networks trained on similar tasks. The remainder of the network - that now has less parameters but has the same complexity - can then be trained for the target task as normal. Training partially frozen networks for Medical Imaging task can mitigate the issues caused by the lack of large-scale annotated datasets.

\subsubsection{Using GANs for Synthetic Data Generation}
\label{sec:useGANs}
Generative Adversarial Networks (GANs)~\cite{goodfellow2014generative} are currently receiving tremendous attention of Computer Vision community for their ability to mimic the distributions from which images are sampled. Among other uses of GANs, one can use the GAN framework to generate realistic synthetic images for any domain.
These images can then be used to train deeper models for that domain that generally outperform the models trained with only (limited) original data. This property of GANs is of particular interest for Medical Image Analysis. Therefore, we can expect to see a large number of future contributions in Medical Imaging that will exploit GANs.
In fact, our literature review already found few initial applications of GANs in medical image analysis~\cite{zhang2017semi}, \cite{calimeri2017biomedical}, \cite{costa2018end}, \cite{lahiri2017generative}.
However, extra care should be taken while exploiting the GAN framework. It should be noted that GANs do not actually learn the original distribution of  images, rather they only mimic it. Hence, the synthetic images generated from GANs can still be very different from the original images. Therefore, instead of training the final model with the data that includes GAN-generated data, it is often better to finally fine-tune such model with only the original images. 

\subsubsection{Miscellaneous Data Augmentation Techniques}
\label{sec:aug}
In general, Computer Vision and Pattern Recognition literature has also developed few elementary data augmentation techniques that have shown improvement in the performance of deep models.
Whereas these techniques are generally not as effective as sophisticated methods, such as using GANs to increase data samples; they are still worth taking advantage of.
We list the most successful techniques below. Again, we note that some of these methods have already proven their effectiveness in the context of Medical Image Analysis:
\begin{itemize}
    \item \textit{Image flipping}: A simple sideways flip of images doubles the number of training samples, that often results in a better model. For medical images, top-down flip is also a possibility due to the nature of images.  
    \item \textit{Image cropping:} Cropping different areas of a larger image into smaller images and treating each one of the cropped versions as an original image also benefits deep models. Five crops of equal dimensions from an image is a popular strategy in Computer Vision literature. The crops are made using the four corners and the central region of the image. \item \textit{Adversarial training:} Very recently, it is discovered that we can `fool' deep models using \textit{adversarial images}~\cite{NaveedSurvey}. These images are carefully computed such that they appear the same as the original images to humans, however, a deep model is not able to recognize them. Whereas developing such images is a different research direction, one finding from that direction is that including those images in training data can improve the performance of deep models~\cite{goodfellow6572explaining}. Since adversarial examples are generated from the original images, they provide a useful data augmentation method that can be harnessed for Medical Imaging tasks.  
    \item \textit{Rotation and random noise addition:} In the context of 3D data, rotating the 3D scans and adding small amount of random noise (emulating jitters) are also considered useful data augmentation strategies~\cite{qi2017pointnet}.
\end{itemize}

\subsection{Enhancing dataset sizes}
\label{sec:Ed}
Whereas the techniques discussed in Section~\ref{sec:SmallData} can alleviate the issues caused by smaller training datasets, the root cause of those problems can only be eliminated by acquiring Deep Learning compatible  large-scale annotated datasets for Medical Image Analysis tasks. Considering that Deep Learning has started outperforming human experts in Medical Image Analysis tasks~\cite{haenssle2018man}, there is a strong need to implement protocols that make medical reports readily convertible to the formats useful for training computational models, especially Deep Learning models. In this context, techniques from the fields of Document Analysis~\cite{bowen2009document} and Natural Language Processing (NLP)~\cite{chowdhury2003natural} can be used to alleviate the extra burden falling on the medical experts due to the implementations of such protocols. 

Besides generating the new data at large-scales that is useful in learning computational models, it is also important to take advantage of the existing medical records for exploiting the current advances in Deep Learning. To handle the large volume of un-organized data (in terms of compatibility with Machine Learning), data mining with humans-in-the-loop~\cite{yu2015lsun} and active learning~\cite{zhu2006semi} can prove beneficial. Advances in Document Analysis and NLP can also be exploited for this task.

\subsection{Broader outlook}
\label{sec:BO}
We can make one important observation regarding Deep Learning research by exploring the literature of different research fields. That is, the advancement in Deep Learning research has often experienced a quantum leap under the breakthroughs provided by different sister fields.
For example, the `residual learning' concept~\cite{he2016deep} that enabled very deep networks was first introduced in the literature of Computer Vision.
This idea (along with other breakthroughs in core Machine Learning research) eventually enabled the tabula rasa algorithm of AlphaGo Zero~\cite{silver2017mastering}. Following up on this observation, we can argue that significant advances can be made in Deep Learning research  in the context of Medical Image Analysis if  researchers from the sister fields of Computer Vision and Machine Learning are able to better understand the Medical Image Analysis tasks. 

Indeed, Medical Imaging community already involves experts from other related fields. However, this involvement is at a smaller scale. For the involvement of  broader Machine Learning and Computer Vision communities, a major hindrance is the Medical Imaging literature jargon. Medical literature is not easily understood by the experts of other fields. One effective method to mitigate this problem can be regular organization of Medical Imaging Workshops and Tutorials in the reputed Computer Vision and Machine Learning Conferences, e.g. IEEE CVPR, ICCV, NeurIPS and ICML. These events should particularly focus on playing the role of translating the Medical Imaging problems to other communities in terms of their topics of interest.      

Another effective strategy to take advantage of Deep Learning advances is to outsource the Medical Imaging problems by organizing online challenges, e.g. \href{https://www.kaggle.com/competitions}{Kaggle competitions}. The authors are already aware of few Kaggle competitions related to Medical Imaging, e.g.~\href{https://www.kaggle.com/c/histopathologic-cancer-detection}{Histopathologic cancer detection}.  However, we can easily notice that Medical Imaging competitions are normally  attracting fewer teams as compared to other competitions - currently 361 for Histopathologic cancer detection. Generally, the number of teams are orders of magnitude lower for the Medical Imaging competitions than those for the  typical imaging competitions. In authors' opinion, strict Medical parlance adopted in organizing such competitions is the source of this problem.
Explanation of Medical Imaging tasks using the terms more common among Computer Vision and Machine Learning communities can greatly help in improving the popularity of Medical Imaging in those communities.

In short, one of the key strategies to fully exploit Deep Learning advances in Medical Imaging is to get the experts from other fields, especially Computer Vision and Machine Learning; to involve in solving Medical Imaging tasks. To that end, the Medical Imaging community must put an extra effort in making its literature, online competitions and the overall outlook of the filed more understandable to the experts from the other fields.
Deep Learning is being dubbed as `modern electricity' by the experts. In the future, its ubiquitous nature will benefit those fields the most that are better understood by the wider communities.

\section{Conclusion}
\label{sec:conc}
This article presented a review of the recent literature in Deep Learning for Medical Imaging. It contributed along three major directions. First, we presented an instructive introduction to the core concepts of Deep Learning. Keeping in view the general lack of understanding of  Deep Learning framework among Medical Imaging  researchers, we kept our discussion intuitive. This part of the paper can be understood as a tutorial of Deep Learning concepts commonly used in Medical Imaging. The second part of the paper presented a comprehensive overview of the approaches in Medical Imaging that employ Deep Learning.
Due the availability of other review articles until the year 2017, we mainly focused on the literature published in the year 2018. The third major part of the article discussed the major challenges faced by Deep Learning in Medical Image Analysis, and discussed the future directions to address those challenges.

Beside focusing on the very recent literature, this article is also different from the existing related literature surveys in that it provides a Computer Vision/Machine Learning perspective to the use of Deep Learning in Medical Image Analysis.
Using that perspective, we are not only able to provide an intuitive understanding of the core  concepts in Deep Learning for the Medical community, we also  highlight the root cause of the challenges faced in this direction and recommend fruitful future directions by drawing on the insights from multiple scientific fields.

From the reviewed literature, we can single out the `lack of large-scale annotated datasets' as the major problem for Deep Learning in Medical Image Analysis. We have discussed and recommended multiple strategies for the Medical Imaging community that are adopted to address similar problems in the sister scientific fields. We can conclude that Medical Imaging can benefit significantly more from Deep Learning by encouraging collaborative research with Computer Vision and Machine Learning research communities.

\bibliographystyle{IEEEtran}
\bibliography{bare_jrnl_compsoc}

\begin{IEEEbiography}[{\includegraphics[width=1in,height=1.25in,clip,keepaspectratio]{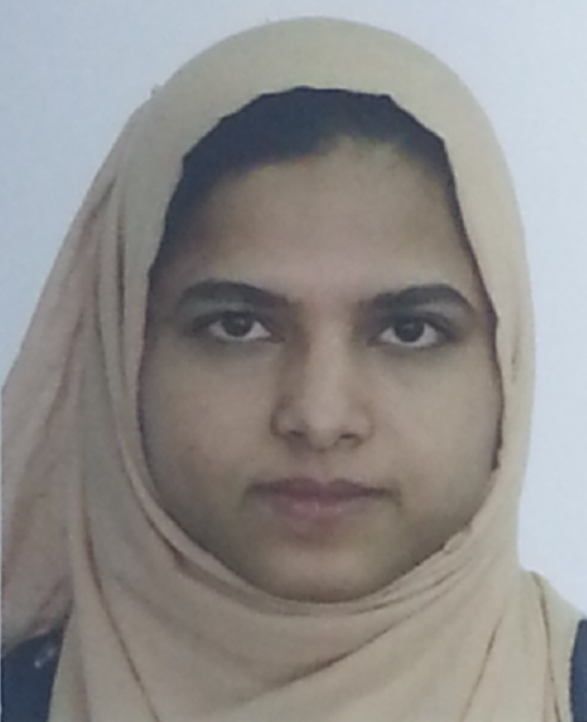}}]{Fouzia Altaf} is currently pursuing her PhD at the School of Science, Edith Cowan University (ECU), Australia. Previously, she completed her Master of Philosophy and Master of Science degrees from the University of the Punjab, Pakistan. She is the recipient of Research Training Program scholarship at ECU. Her current research pertains to the areas of Deep Learning, Medical Image Analysis and Computer Vision.
\end{IEEEbiography}

\begin{IEEEbiography}[{\includegraphics[width=1in,height=1.25in,clip,keepaspectratio]{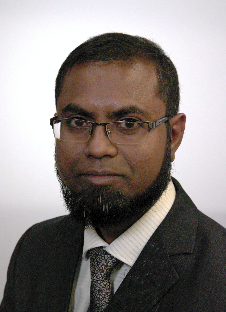}}]{Syed Islam} completed his PhD with Distinction in Computer Engineering from the University of Western Australia (UWA) in 2011. He received his MSc in Computer Engineering from King Fahd University of Petroleum and Minerals in 2005 and BSc in Electrical and Electronic Engineering from Islamic Institute of Technology in 2000. He was a Research Assistant Professor at UWA from 2011 to 2015, a Research Fellow at Curtin University from 2013 to 2015 and a Lecturer at UWA from 2015-2016. Since 2016, he has been working as Lecturer in Computer Science at Edith Cowan University. He has published around 53 research articles and got nine public media releases. He obtained 14 competitive research grants for his research in the area of Image Processing, Computer Vision and Medical Imaging. He has co-supervised to completion seven honours and postgrad students and currently supervising one MS and seven PhD students. He is serving as Associate Editor of 13 international journals, Technical Committee Member of 24 conferences and regular reviewer of 26 journals. He is also serving seven professional bodies including IEEE and Australian Computer Society (Senior Member).

\end{IEEEbiography}

\begin{IEEEbiography}[{\includegraphics[width=1in,height=1.25in,clip,keepaspectratio]{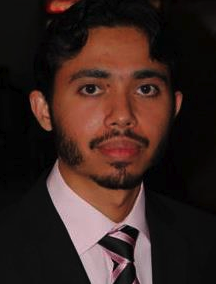}}]{Naveed Akhtar} received his PhD in Computer Vision from The University of Western Australia (UWA) and Master degree in Computer Science from Hochschule Bonn-Rhein-Sieg, Germany (HBRS). His research in  Computer Vision and Pattern Recognition is regularly published in the  prestigious venues of the field, including IEEE CVPR and IEEE TPAMI. He has also served as a reviewer for these, and twelve other research venues in the broader field of Machine Learning. He is currently also serving as an Associate Editor of the IEEE Access. During his PhD, he was a recipient of multiple scholarships, and winner of the Canon Extreme Imaging Competition in 2015. 
Currently, he is a Research Fellow at UWA. Previously, he has also served on the same position at the Australian National University. His current research interests include Medical Imaging, Adversarial Machine Learning, Hyperspectral Image Analysis, and 3D Point Cloud Analysis.
\end{IEEEbiography}

\begin{IEEEbiography}[{\includegraphics[width=1in,height=1.25in,clip,keepaspectratio]{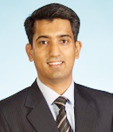}}]{Naeem Khalid Janjua} is a Lecturer at the School of Science, Edith Cowan University, Perth.  He is an Associate Editor for International Journal of Computer System Science and Engineering (IJCSSE) and International Journal of Intelligent systems (IJEIS). He has published an authored book, a book chapter, and various articles in international journals and refereed conference proceedings. His areas of active research are defeasible reasoning, argumentation, ontologies, data modelling, cloud computing, machine learning and data mining. He works actively in the domain of business intelligence and Web-based intelligent decision support systems.

\end{IEEEbiography}


\end{document}